\documentclass[11pt]{article}

\usepackage[preprint]{acl}

\usepackage{times}
\usepackage{latexsym}
\usepackage{amsmath}
\usepackage{amssymb}
\usepackage{placeins}
\usepackage[T1]{fontenc}

\usepackage[utf8]{inputenc}

\usepackage{microtype}

\usepackage{inconsolata}
\usepackage{graphicx}
\usepackage{booktabs}
\usepackage{multirow}
\usepackage{tabularx}
\usepackage{enumitem}
\usepackage[ruled,vlined,linesnumbered]{algorithm2e}
\usepackage[most]{tcolorbox}
\usepackage[capitalise]{cleveref}
\usepackage{pgfplots}
\pgfplotsset{compat=1.18}

\definecolor{problue}{HTML}{0072B2}
\definecolor{outred}{HTML}{D55E00}
\definecolor{searchorange}{HTML}{E69F00}
\definecolor{steppurple}{HTML}{CC79A7}
\definecolor{reasongreen}{HTML}{009E73}
\usepackage[most]{tcolorbox}
\usepackage{xcolor}

\definecolor{policyblue}{HTML}{2F6F9F}
\definecolor{labelgreen}{HTML}{3D7A57}
\definecolor{prmorange}{HTML}{B56A32}

\newtcolorbox{promptbox}[3][]{
    enhanced,
    breakable,
    colback=#2!4,
    colframe=#2,
    boxrule=0.8pt,
    arc=1.5mm,
    left=3mm,
    right=3mm,
    top=2.5mm,
    bottom=2.5mm,
    fontupper=\small,
    fonttitle=\bfseries,
    coltitle=white,
    colbacktitle=#2,
    title={#3},
    attach boxed title to top left={
        xshift=3mm,
        yshift=-2mm
    },
    boxed title style={
        boxrule=0pt,
        arc=1mm,
        left=2mm,
        right=2mm,
        top=1mm,
        bottom=1mm
    },
    before skip=2mm,
    after skip=2mm,
    #1
}
\title{PRO-Step: Step-level Process Reward Optimization for Retrieval-Augmented Generation}

\author{
MinKeon Kim \textbullet\ Namjun Lee \textbullet\
Jaekwang Kim\textsuperscript{*}\\
Department of Applied Artificial Intelligence,\\
Convergence Program for Social Innovation,\\
Sungkyunkwan University, Seoul, South Korea\\
\texttt{alsrjs0418@skku.edu},
\texttt{namejun12000@g.skku.edu},
\texttt{linux@skku.edu}
}

\begin{document}
\maketitle
\begingroup
\renewcommand{\thefootnote}{\fnsymbol{footnote}}
\setcounter{footnote}{1}
\footnotetext{Corresponding author.}
\endgroup

\begin{abstract}
Retrieval-Augmented Generation enhances Large Language Models by grounding responses in external knowledge, but multi-hop reasoning remains vulnerable to error propagation, where early retrieval failures confound subsequent steps. Standard outcome-based optimization only rewards the final answer, leaving intermediate retrieval and reasoning errors undetected. While existing process-based methods introduce step-level signals, they still score each step against the final answer, rewarding spurious successes where flawed retrieval coincidentally produces the correct answer. Step-level supervision in RAG requires evaluating both logical validity and evidential grounding at each step. We introduce \textsc{PRO-Step}: we train a generative PRM that evaluates both dimensions, employ PRM-guided value tree search to construct preference pairs contrasting valid steps against flawed ones, and optimize the policy via step-level Direct Preference Optimization. Experiments on single- and multi-hop QA datasets demonstrate that \textsc{PRO-Step} achieves the best average EM and F1 across five benchmarks.
Code, models, and training data are publicly available at
\href{https://github.com/keemminnke/PRO-Step}{https://github.com/keemminnke/PRO-Step}.
\end{abstract}

\section{Introduction}
\label{sec:introduction}

Large Language Models (LLMs) \citep{gpt5systemcard, deepseekr1} have demonstrated capabilities across a wide range of reasoning tasks, including mathematics and code generation. However, they still suffer from hallucinations, often generating plausible but unsupported claims when they lack relevant or up-to-date knowledge. Retrieval-Augmented Generation (RAG) \citep{lewis2020rag} addresses these issues by equipping LLMs with external knowledge sources, improving factual accuracy and timeliness. However, traditional RAG follows a linear retrieve-then-generate workflow. This design struggles with multi-step reasoning tasks that require iterative retrieval and dynamic query reformulation. To overcome this, recent work has proposed agentic RAG systems \citep{search-o1,research2025}, enabling models to dynamically decide when and what to retrieve through iterative retrieval, query rewriting, and self-reflection mechanisms \citep{jiang2023flare,su2024dragin,asai2024selfrag}.

Early approaches \citep{research2025} relied on prompt engineering, while Supervised Fine-Tuning (SFT) methods directly optimized model parameters for retrieval-augmented workflows. More recently, outcome-supervised Reinforcement Learning (RL) \citep{ouyang2022training} has achieved improvements: Search-R1 \citep{searchr1} and R1-Searcher \citep{r1searcher2025} incorporate search engines into the LLM's environment and use final-answer correctness as the reward signal, demonstrating that end-to-end RL can enhance agentic RAG capabilities.

However, outcome-based optimization presents limitations in multi-hop reasoning \citep{zhang2025reasonrag}. First, a minor retrieval error in the early steps, such as an imprecise query or an irrelevant document, propagates through subsequent reasoning steps, but outcome-based optimization lacks the means to detect or penalize these intermediate failures. Second, because the reward signal is derived solely from final-answer correctness, it is sparse. The model receives feedback only after completing the entire reasoning chain, requiring more training data to converge.

Process Reward Model (PRM) approaches \citep{lightman2023letsverify,omegaprm2024,mathshepherd2024,zheng-etal-2025-processbench} address this limitation by providing step-level feedback that guides models along correct reasoning trajectories. However, applying existing PRM approaches to RAG presents two challenges. First, they focus on logical consistency in domains like mathematics and cannot verify whether a claim is factually grounded in retrieved documents. Second, most PRM approaches produce only scalar scores without explaining why a step is flawed, limiting the richness of the supervision signal.

To address these challenges, we propose \textsc{PRO-Step}, a framework that integrates process-level supervision into RAG optimization. \textsc{PRO-Step} consists of two stages: (1) \textbf{Generative PRM training.} We generate diverse Chain-of-Thought trajectories where reasoning and retrieval are interleaved, and train a Generative PRM that evaluates each step with a correctness label and a rationale. (2) \textbf{Policy optimization.} We construct step-level preference pairs via PRM-guided value
tree search (VTS), and apply Direct Preference Optimization (DPO) to train the policy model to reason correctly at each step.
Our contributions are as follows:

\begin{itemize}
\item We identify that existing process-level RAG methods still optimize against final-answer outcomes, failing to penalize spurious reasoning because they lack mechanisms to evaluate both logical validity and evidential grounding at each step.
\item We propose \textsc{PRO-Step}, a framework that combines Generative PRM training, PRM-guided VTS for preference pair construction, and process-supervised DPO. By providing dense step-level feedback, this approach mitigates the reward sparsity issue inherent in outcome-based methods, enabling efficient optimization for retrieval-augmented reasoning.
\item Through experiments on single-hop and multi-hop QA datasets, we demonstrate that \textsc{PRO-Step} achieves the best average EM and F1 among the evaluated methods across five QA benchmarks.
\end{itemize}

\section{Related Work}
\label{sec:related-work}

\begin{figure*}[t]
\centering
\vspace{-2mm}
\includegraphics[width=1.0\textwidth,trim=0 0 0 0,clip]{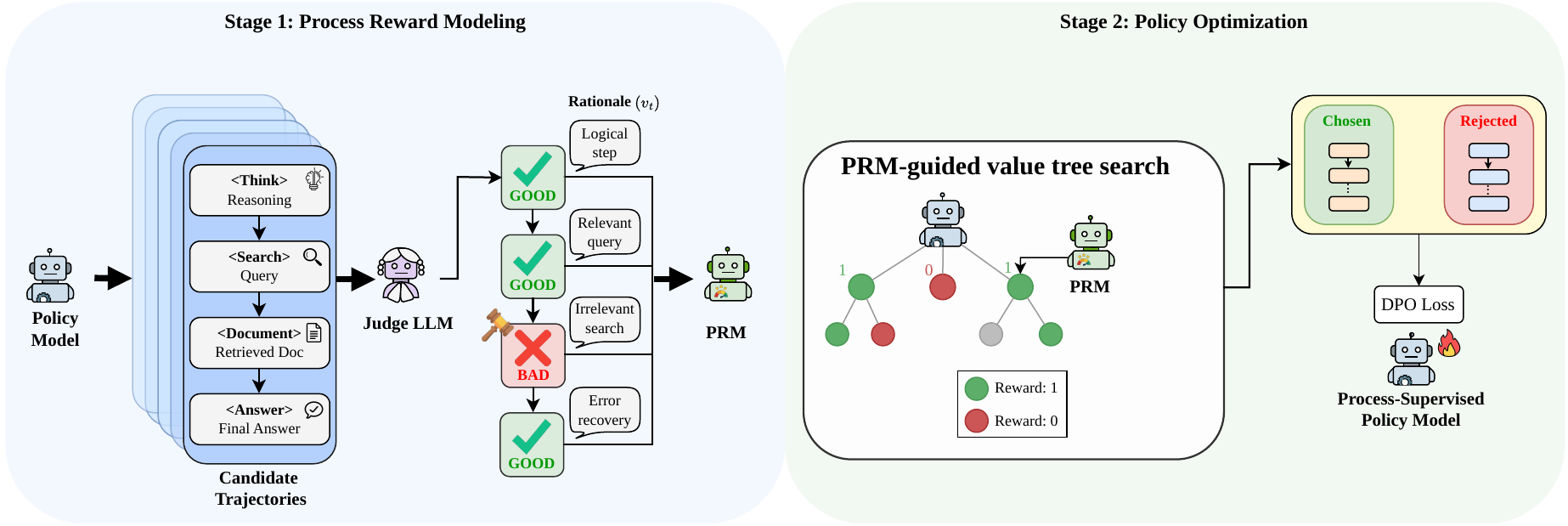}
\vspace{-3mm}
\caption{\textbf{\textsc{PRO-Step} Overview:} Detailed architecture of the process-supervised optimization framework.}
\label{fig:overview}
\vspace{-4mm}
\end{figure*}

\subsection{Retrieval-Augmented Generation}
Early RAG systems followed a retrieve-then-generate paradigm, appending fixed passages to the input prompt \citep{asai2024selfrag}. While effective for single-hop queries, this approach struggles with multi-hop scenarios necessitating iterative retrieval and reasoning. Prompt-based methods introduced uncertainty-triggered retrieval (FLARE; \citealt{jiang2023flare}), dynamic query reformulation (DRAGIN; \citealt{su2024dragin}), and interleaved reasoning-retrieval chains (\citealt{trivedi2022ircot}). RL-based frameworks subsequently enabled autonomous retrieval decisions: Search-R1, R1-Searcher, and related systems incorporate search environments using final-answer correctness as the reward, showing that end-to-end outcome-based RL improves agentic retrieval performance \citep{research2025}. Despite this success, outcome-only rewards remain sparse and provide no signal on intermediate retrieval or reasoning quality, motivating recent shifts toward process-level supervision.

Process-level supervision has been applied to RAG through step-level rewards \citep{r3rag2025, prorag2026, zheng2025stepsearch, hiprag2025} and tree search \citep{grat2025, zhang2025reasonrag}. However, these methods still score intermediate steps against final-answer correctness, failing to filter out spurious successes where flawed retrieval coincidentally produces the correct answer.

\subsection{Process Reward Model Approaches}
PRM approaches provide step-level verification for multi-step reasoning. \citet{lightman2023letsverify} first demonstrated that PRM-based verification improves mathematical reasoning by independently verifying steps. Subsequent works scaled PRM training using automated Monte Carlo rollouts (OmegaPRM; \citealt{omegaprm2024}) and step-level outcome estimation (Math-Shepherd; \citealt{mathshepherd2024}).

Moving beyond scalar evaluation, GenPRM \citep{genprm2026} introduced a generative paradigm where the model articulates rationales before assigning a correctness label. By generating these rationales, the PRM not only scores steps but explicitly identifies errors, providing richer feedback for policy optimization.

However, existing PRM research remains concentrated on formal reasoning, where step correctness is logically verifiable. RAG environments present a different challenge: each step interleaves internal reasoning with external retrieval, requiring verification of both logical validity and evidential grounding in retrieved passages. While VersaPRM \citep{versaprm2025} extends PRM approaches to multiple domains, it does not address the dependency between retrieval and reasoning. \textsc{PRO-Step} addresses this gap by training a Generative PRM that evaluates both dimensions, providing step-level signals that guide value tree search for preference pair construction and policy optimization.

\section{\textsc{PRO-Step}}
\label{sec:method}

We introduce \textsc{PRO-Step}, a two-stage framework.
Section \ref{sec:psdg} describes the construction of process-supervised data, including trajectory generation, step annotation, and generative PRM training.
Section \ref{sec:policy-opt} presents policy optimization using a PRM-guided value tree search and step-level DPO.

\begin{figure*}[!t]
\centering
\vspace{-2mm}
\includegraphics[width=1.0\textwidth,trim=0 0 0 0,clip]{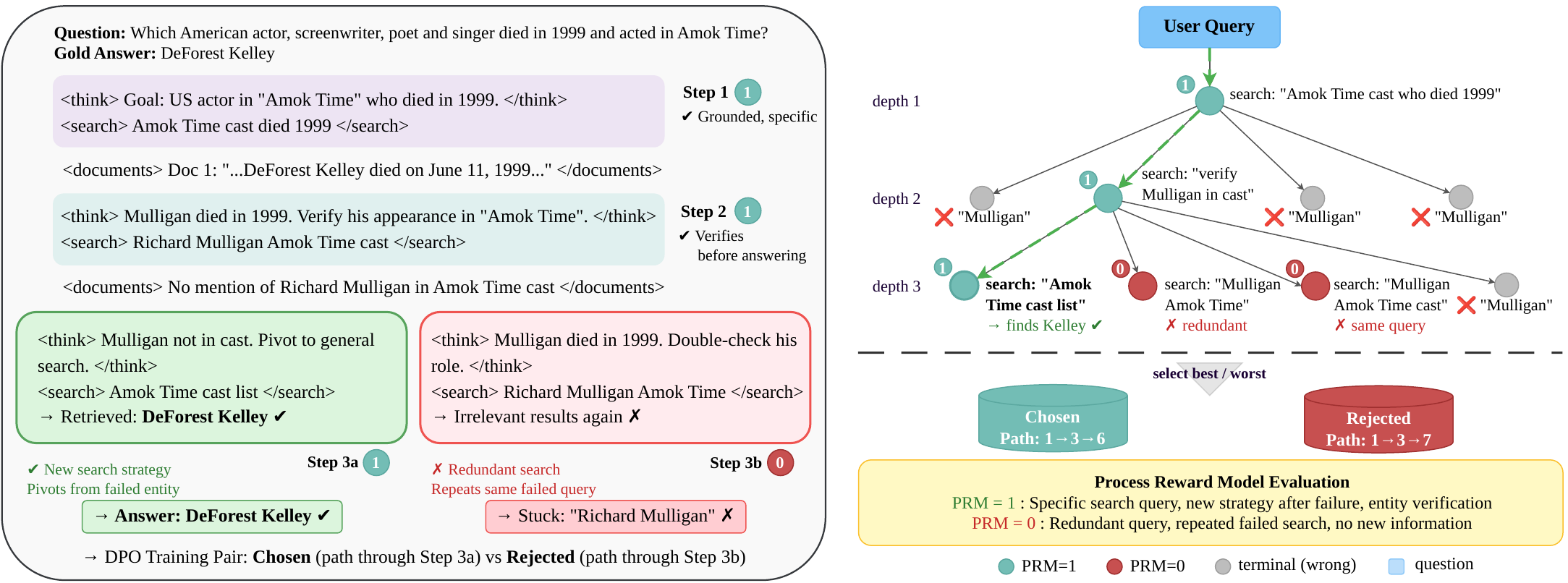}
\vspace{-3mm}
\caption{Example of tree-search data generation with a PRM. Left: reasoning trace showing PRM evaluation at each step. Right: search tree where green nodes (PRM=1) indicate sound reasoning and red nodes (PRM=0) indicate flawed steps. The best-scoring path becomes chosen and the worst becomes rejected for DPO training.}
\label{fig:mcts_prm_generation}
\vspace{-4mm}
\end{figure*}

\subsection{Process Reward Modeling}
\label{sec:psdg}

Although existing PRM approaches evaluate logical steps in mathematical reasoning, they lack mechanisms for factual grounding in external evidence, failing to outperform majority voting in RAG contexts (\Cref{fig:prm_comparison}). To address this limitation, we construct a process-supervised dataset by generating reasoning trajectories, annotating step-level labels, and training a generative PRM specific to RAG.

\paragraph{Trajectory generation.}
We formulate trajectory generation as a multi-turn interaction between the policy model $\pi_\theta$ (using the prompt in \Cref{app:Prompt}) and an external search engine $\mathcal{E}$. $\pi_\theta$ generates its internal reasoning within \texttt{<think>...</think>}. When external factual grounding is necessary, the model triggers retrieval by generating \texttt{<search>...</search>}. Upon detecting a search request, the system halts generation, queries $\mathcal{E}$ with the generated search string, and appends the retrieved documents back into the sequence as \texttt{<documents>...</documents>}. This interleaved generation-and-retrieval cycle continues until the model outputs a final prediction formatted as \texttt{<answer>...</answer>} or exhausts the maximum interaction steps. Using this procedure, we sample 16 candidate reasoning trajectories per question from 2,000 instances drawn from HotpotQA \citep{yang2018hotpotqa} and MuSiQue \citep{trivedi2022musique}. After discarding incomplete paths or those lacking formatted answers, we retain 31,728 trajectories.

\paragraph{Step annotation.}
We use QwQ-32B \citep{qwq2024} to annotate each model-generated step $s_t$ based on the evaluation prompt provided in \Cref{app:Prompt}. Given $q$ and the preceding context $s_{\le t}$, QwQ-32B generates a rationale $v_t$ and a binary correctness label $r_t \in \{0, 1\}$. Unlike prior PRM methods that discard all subsequent steps once an error occurs, we retain the entire trajectory. In RAG contexts, because $\pi_\theta$ can recover from an incorrect step by issuing a revised search query, a local failure ($r_t = 0$) does not invalidate subsequent steps. All 31,728 trajectories are annotated to form $\mathcal{D}_{\text{PRM}}$ (dataset statistics in \Cref{app:data-stats}, labeling rubric in \Cref{app:labeling_rubric}, and annotation quality audit in \Cref{app:prm_validation}).

\paragraph{Generative PRM training.}
We train a generative process reward model $\pi_\psi$ (initialized from DeepSeek-R1-Distill-8B \citep{deepseekr1, qwen3}) on $\mathcal{D}_{\text{PRM}}$. The model is optimized to generate both the rationale $v_t$ and the binary correctness label $r_t$ by minimizing the following objective:
\begin{equation}
\label{eq:prm-loss}
\begin{aligned}
\mathcal{L}_{\text{PRM}} &= -\mathbb{E}_{(q,\tau,v_t,r_t) \sim \mathcal{D}_{\text{PRM}}}\Bigl[
\log \pi_\psi(v_t \mid q, s_{\le t}) \\
&\qquad + \log \pi_\psi(r_t \mid q, s_{\le t}, v_t)
\Bigr]
\end{aligned}
\end{equation}
where the expectation is computed over the model-generated steps. At inference, $\pi_\psi$ first generates the rationale, and then outputs the binary prediction:
\begin{equation}
\label{eq:prm-inference}
\hat{r}(s_t) = \arg\max_{r \in \{0,1\}} \pi_\psi(r \mid q, s_{\le t}, v_t)
\end{equation}

\subsection{Policy Optimization}
\label{sec:policy-opt} 
Standard process-level methods evaluate intermediate steps by simulating trajectories to completion and scoring the final outcome. However, this approach lacks mechanisms to penalize flawed reasoning that spuriously reaches the correct answer.
Unlike prior process-based methods that still rely heavily on final outcomes, \textsc{PRO-Step} actively filters out these spurious successes by using a PRM-guided value tree search to explicitly penalize flawed intermediate steps.

\paragraph{PRM-guided Value Tree Search.}

Following modern tree search paradigms such as AlphaGo Zero \citep{silver2017mastering}, we adopt a value tree search framework that explicitly replaces standard random rollouts with direct value estimation. Within this framework, each node represents a state $s_t = (q, s_{\le t})$, and an action corresponds to sampling the next model-generated step from $\pi_\theta$. If the sampled step is a \texttt{<search>} action, $\mathcal{E}$ appends the corresponding \texttt{<documents>} block before the next action is taken. The tree is built via iterating selection, expansion, and evaluation using upper confidence bound (UCB)-based selection \citep{kocsis2006bandit} 
and back-propagation (detailed in Appendix \ref{app:VTS}).

For a terminal node, the reward is the depth-discounted F1 score $Q(s_T)$. For a non-terminal node $s_t$, we directly use the binary prediction from $\pi_\psi$:
\begin{equation}
\label{eq:q-nonterminal}
Q(s_t) = \hat{r}(s_t) \cdot \gamma^{d(s_t)}
\end{equation}
where $\gamma \in (0,1)$ penalizes long trajectories, and $d(\cdot)$ denotes the depth of the node. The prediction $\hat{r}(s_t)$ is computed after node expansion and propagated to all ancestor nodes, updating the expected value $\bar{Q}(s)$ without inflating visit counts. Detailed VTS hyperparameters and a branching-factor ablation are
provided in \Cref{app:VTS}.

\paragraph{Preference data construction.}
For each parent node, we rank its child nodes based on a value function 
that balances the outcome quality with the assessment from $\pi_\psi$:
\begin{equation}
\label{eq:value-composite}
V(s) = \bar{Q}(s) + \alpha \cdot \hat{r}(s)
\end{equation}
Here, $\bar{Q}(s)$ aggregates the quality of descendant trajectories 
through back-propagation, while $\hat{r}(s) \in \{0, 1\}$ assesses the 
correctness of step $s$. For any sibling pair $s_i$ and $s_j$ where $V(s_i) - V(s_j) > \delta$, we assign the higher-valued node as the chosen step ($s^+ = s_i$) and the other as the rejected step ($s^- = s_j$).

\paragraph{DPO training.}
We optimize $\pi_\theta$ using step-level DPO \citep{rafailov2023dpo}. First, we define the implicit utility $u_\theta(x, y_t)$:
\begin{equation}
\label{eq:implicit-utility}
u_\theta(x, y_t) = \beta \log \frac{\pi_\theta(y_t \mid x, y_{<t})}{\pi_{\text{ref}}(y_t \mid x, y_{<t})}
\end{equation}
Using this utility, we minimize the DPO loss, masking \texttt{<documents>} blocks from the computation:
\begin{equation}
\label{eq:dpo-loss-final}
\begin{aligned}
\mathcal{L}_{\text{DPO}}(\theta)
&= -\mathbb{E}_{\mathcal{D}_{\text{dpo}}}\Bigl[
\log \sigma\!\left(u_\theta(x, y_t^w)\right. \\
&\qquad\left. - u_\theta(x, y_t^l)\right)\Bigr]
\end{aligned}
\end{equation}
where $\mathcal{D}_{\text{dpo}}$ is the step-level preference dataset of tuples $(x, y_{<t}, y_t^w, y_t^l)$. Here, $x$ denotes the question with system prompt, $y_{<t}$ is the shared reasoning prefix, $y_t^w$ and $y_t^l$ are the chosen and rejected steps at depth $t$, and the hyperparameter $\beta$ controls the KL constraint.

\begin{table*}[!t]
\centering
\caption{Comparison of different RAG methods across 5 benchmarks using EM and F1 metrics. \textbf{Bold} denotes the best performance, and \underline{underlining} denotes the second-best. $^{\dagger}$ indicates statistically significant improvements ($p < 0.05$) over representative RL baselines (Search-R1, ReasonRAG) using a two-sided paired $t$-test. Full significance tests, including comparisons with StepSearch, are detailed in Appendix \ref{app:significance}.}

\label{tab:main_results}
\resizebox{\textwidth}{!}{%
\renewcommand{\arraystretch}{1.10}
\begin{tabular}{@{}llcccccccccc|cc@{}}
\toprule
\multirow{2}{*}{\textbf{Type}} & \multirow{2}{*}{\textbf{Method}} & \multicolumn{2}{c}{\textbf{PopQA}} & \multicolumn{2}{c}{\textbf{HotpotQA}} & \multicolumn{2}{c}{\textbf{2WikiMulti}} & \multicolumn{2}{c}{\textbf{Bamboogle}} & \multicolumn{2}{c|}{\textbf{MuSiQue}} & \multicolumn{2}{c}{\textbf{Average}} \\
\cmidrule(lr){3-4} \cmidrule(lr){5-6} \cmidrule(lr){7-8} \cmidrule(lr){9-10} \cmidrule(lr){11-12} \cmidrule(lr){13-14}
& & EM & F1 & EM & F1 & EM & F1 & EM & F1 & EM & F1 & EM & F1 \\
\midrule
\multirow{2}{*}{Zero-shot} & Direct & 14.8 & 19.0 & 18.7 & 26.6 & 24.6 & 29.3 & 12.0 & 19.2 & 4.0 & 11.9 & 14.8 & 21.2 \\
& Standard RAG & 37.4 & 44.1 & 31.6 & 41.9 & 28.3 & 34.6 & 16.8 & 25.1 & 4.8 & 12.3 & 23.8 & 31.6 \\
\midrule
\multirow{4}{*}{Advanced} & FLARE & 15.5 & 19.6 & 18.8 & 26.6 & 24.6 & 29.0 & 12.0 & 19.2 & 4.0 & 11.6 & 15.0 & 21.2 \\
& Self-RAG & 30.0 & 38.6 & 18.2 & 29.6 & 13.2 & 25.0 & 5.6 & 18.0 & 4.7 & 11.8 & 14.3 & 24.6 \\
& IRCoT & 32.7 & 40.7 & 31.2 & 42.8 & 26.2 & 35.1 & 23.2 & 33.6 & 6.2 & 13.0 & 23.9 & 33.0 \\
& Search-o1 & 10.7 & 15.0 & 18.5 & 26.5 & 19.6 & 23.4 & 25.6 & 38.5 & 4.5 & 11.8 & 15.8 & 23.0 \\
\midrule
\multirow{4}{*}{RL-based} & AutoRAG & 24.1 & 31.2 & 14.8 & 24.1 & 11.0 & 20.4 & 23.2 & 34.3 & 6.5 & 13.1 & 15.9 & 24.6 \\
& Search-R1 & \textbf{40.7} & \underline{46.8} & 37.9 & 49.6 & 34.9 & 42.5 & 33.6 & 43.6 & \underline{13.0} & 21.2 & 32.0 & 40.7 \\
& ReasonRAG & 37.8 & 44.9 & 36.4 & 47.5 & 39.8 & 46.3 & \textbf{38.4} & \underline{46.9} & 10.6 & 19.2 & 32.6 & 41.0 \\
& StepSearch & 39.2 & 45.0 & \textbf{38.7} & \underline{50.7} & \underline{40.4} & \underline{47.1} & 33.6 & 44.2 & \textbf{13.8} & \textbf{23.1} & \underline{33.2} & \underline{42.0} \\
\midrule
\textbf{Ours} & \textbf{\textsc{PRO-Step}} & \underline{40.5} & \textbf{47.4}$^{\dagger}$ & \textbf{38.7} & \textbf{51.6}$^{\dagger}$ & \textbf{44.1}$^{\dagger}$ & \textbf{51.4}$^{\dagger}$ & \underline{36.8} & \textbf{47.6} & 12.5 & \underline{22.4} & \textbf{34.5}$^{\dagger}$ & \textbf{44.1}$^{\dagger}$ \\
\bottomrule
\end{tabular}%
}
\end{table*}
\section{Experiments}
\label{sec:experiments}
In this section, we present the experimental settings and results.

\begin{table*}[!t]
\centering
\caption{Component ablation study evaluating the impact of the Process Reward Model (PRM). The table reports EM and F1 scores across 5 benchmarks. \textbf{Bold} denotes the better performance.}
\label{tab:component_ablation}
\resizebox{\textwidth}{!}{%
\renewcommand{\arraystretch}{1.10}
\begin{tabular}{@{}lcccccccccccc@{}}
\toprule
\multirow{2}{*}{\textbf{Variant}} & \multicolumn{2}{c}{\textbf{PopQA}} & \multicolumn{2}{c}{\textbf{HotpotQA}} & \multicolumn{2}{c}{\textbf{2WikiMulti}} & \multicolumn{2}{c}{\textbf{Bamboogle}} & \multicolumn{2}{c}{\textbf{MuSiQue}} & \multicolumn{2}{c}{\textbf{Average}} \\
\cmidrule(lr){2-3} \cmidrule(lr){4-5} \cmidrule(lr){6-7} \cmidrule(lr){8-9} \cmidrule(lr){10-11} \cmidrule(lr){12-13}
& EM & F1 & EM & F1 & EM & F1 & EM & F1 & EM & F1 & EM & F1 \\
\midrule
\textbf{\textsc{PRO-Step}} & \textbf{40.5} & \textbf{47.4} & \textbf{38.7} & \textbf{51.6} & \textbf{44.1} & \textbf{51.4} & \textbf{36.8} & \textbf{47.6} & \textbf{12.5} & \textbf{22.4} & \textbf{34.5} & \textbf{44.1} \\
Without PRM & 38.5 & 45.0 & 36.8 & 49.3 & 41.4 & 48.4 & 34.4 & 44.2 & 12.3 & \textbf{22.4} & 32.7 & 41.9 \\
\bottomrule
\end{tabular}%
}
\end{table*}

\begin{table*}[!t]
\centering
\caption{Impact of different optimization strategies on \textsc{PRO-Step}'s effectiveness across five benchmarks. \textbf{Bold} denotes the best performance, and \underline{underlining} denotes the second-best.}
\label{tab:ablation_results}
\resizebox{\textwidth}{!}{%
\renewcommand{\arraystretch}{1.10}
\begin{tabular}{@{}lcccccccccc|cc@{}}
\toprule
\multirow{2}{*}{\textbf{Method}} & \multicolumn{2}{c}{\textbf{PopQA}} & \multicolumn{2}{c}{\textbf{HotpotQA}} & \multicolumn{2}{c}{\textbf{2WikiMulti}} & \multicolumn{2}{c}{\textbf{Bamboogle}} & \multicolumn{2}{c|}{\textbf{MuSiQue}} & \multicolumn{2}{c}{\textbf{Average}} \\
\cmidrule(lr){2-3} \cmidrule(lr){4-5} \cmidrule(lr){6-7} \cmidrule(lr){8-9} \cmidrule(lr){10-11} \cmidrule(lr){12-13}
& EM & F1 & EM & F1 & EM & F1 & EM & F1 & EM & F1 & EM & F1 \\
\midrule
\textsc{PRO-Step} (SFT) & 38.8 & 45.7 & \underline{36.7} & \underline{48.3} & 32.9 & 41.3 & \underline{31.2} & \underline{40.4} & \underline{11.3} & \underline{20.0} & \underline{30.2} & \underline{39.1} \\
\textsc{PRO-Step} (KTO) & \textbf{41.8} & \underline{46.3} & 32.3 & 43.5 & \underline{36.1} & \underline{42.0} & 20.8 & 32.7 & 8.9 & 18.5 & 28.0 & 36.6 \\
\midrule
\textbf{\textsc{PRO-Step} (DPO)} & \underline{40.5} & \textbf{47.4} & \textbf{38.7} & \textbf{51.6} & \textbf{44.1} & \textbf{51.4} & \textbf{36.8} & \textbf{47.6} & \textbf{12.5} & \textbf{22.4} & \textbf{34.5} & \textbf{44.1} \\
\bottomrule
\end{tabular}%
}
\end{table*}

\subsection{Experimental Setup}

\paragraph{Datasets and Evaluation Metrics.}
We evaluate \textsc{PRO-Step} on five benchmark datasets spanning single-hop and multi-hop question answering (QA). For single-hop QA, we use PopQA \citep{mallen2022popqa}. For multi-hop QA, we use HotpotQA \citep{yang2018hotpotqa}, 2WikiMultiHopQA \citep{ho2020constructing}, Bamboogle \citep{press2023measuring}, and MuSiQue \citep{trivedi2022musique}. We report Exact Match (EM) and F1 score using the FlashRAG evaluation framework \citep{FlashRAG} for reproducibility.

\paragraph{Implementation Details.}
We employ Qwen2.5-7B-Instruct \citep{qwen2.5} as the primary backbone for \textsc{PRO-Step} and most baselines. For specific baselines, we utilize their official model checkpoints to ensure a fair comparison: Self-RAG employs its specialized Llama-2-13B variant \citep{asai2024selfrag}, and AutoRAG utilizes Llama-3-8B-Instruct \citep{yu2024autorag}.
For retrieval, we use the 2018 Wikipedia dump consisting of 5.9M documents, indexed with the BGE-base-en-v1.5 \citep{xiao2023bge} retriever, and set the number of retrieved passages to $k=3$ across all methods to ensure consistency.
We train \textsc{PRO-Step} on a curated mixture of 5,000 multi-hop QA questions, comprising 2,000 instances each from the \textit{training splits} of HotpotQA and MuSiQue, and 1,000 from 2WikiMultiHopQA. For PRM training, we sample 16 trajectories for each of the 2,000 questions from Section 3.1 above. For policy optimization, we then construct step-level preference pairs via PRM-guided VTS. Detailed data statistics and filtering procedures are deferred to \Cref{app:data-stats}.

\paragraph{Baselines.}
We evaluate \textsc{PRO-Step} against 10 baselines categorized into three groups: \textbf{Zero-shot Baselines} (Naive Generation, Standard RAG \citep{lewis2020rag}), \textbf{Advanced RAG} featuring adaptive retrieval mechanisms and iterative reasoning capabilities (FLARE \citep{jiang2023flare}, Self-RAG, IRCoT \citep{trivedi2022ircot}, Search-o1 \citep{search-o1}), and \textbf{RL-based RAG} paradigms that leverage autonomous exploration and test-time scaling, including \textit{outcome-reward} methods (AutoRAG, Search-R1 \citep{searchr1}) and \textit{process-reward} methods (ReasonRAG \citep{zhang2025reasonrag}, StepSearch \citep{zheng2025stepsearch}). See \Cref{app:implementation} for training hyperparameters.

\subsection{Main Results}
The main results comparing \textsc{PRO-Step} with baseline methods across five datasets are presented in Table \ref{tab:main_results}. From the results, we make the following key observations:

\paragraph{Overall results.}
\textsc{PRO-Step} achieves the highest average performance across all evaluated methods, recording 34.5 EM and 44.1 F1 (significance tests in \Cref{app:significance}). 

\paragraph{Limitations of heuristic adaptive retrieval.} 
Advanced RAG methods relying on heuristic triggers, such as FLARE and Self-RAG, underperform the much simpler Standard RAG on average (approximately 15.0 vs. 23.8 EM). Even IRCoT falls behind Standard RAG on complex multi-hop tasks such as HotpotQA and 2WikiMultiHopQA. In contrast, \textsc{PRO-Step} learns retrieval timing directly via preference optimization, bypassing the inefficiencies of heuristics.

\begin{figure*}[!t]
\centering
\includegraphics[width=\textwidth]{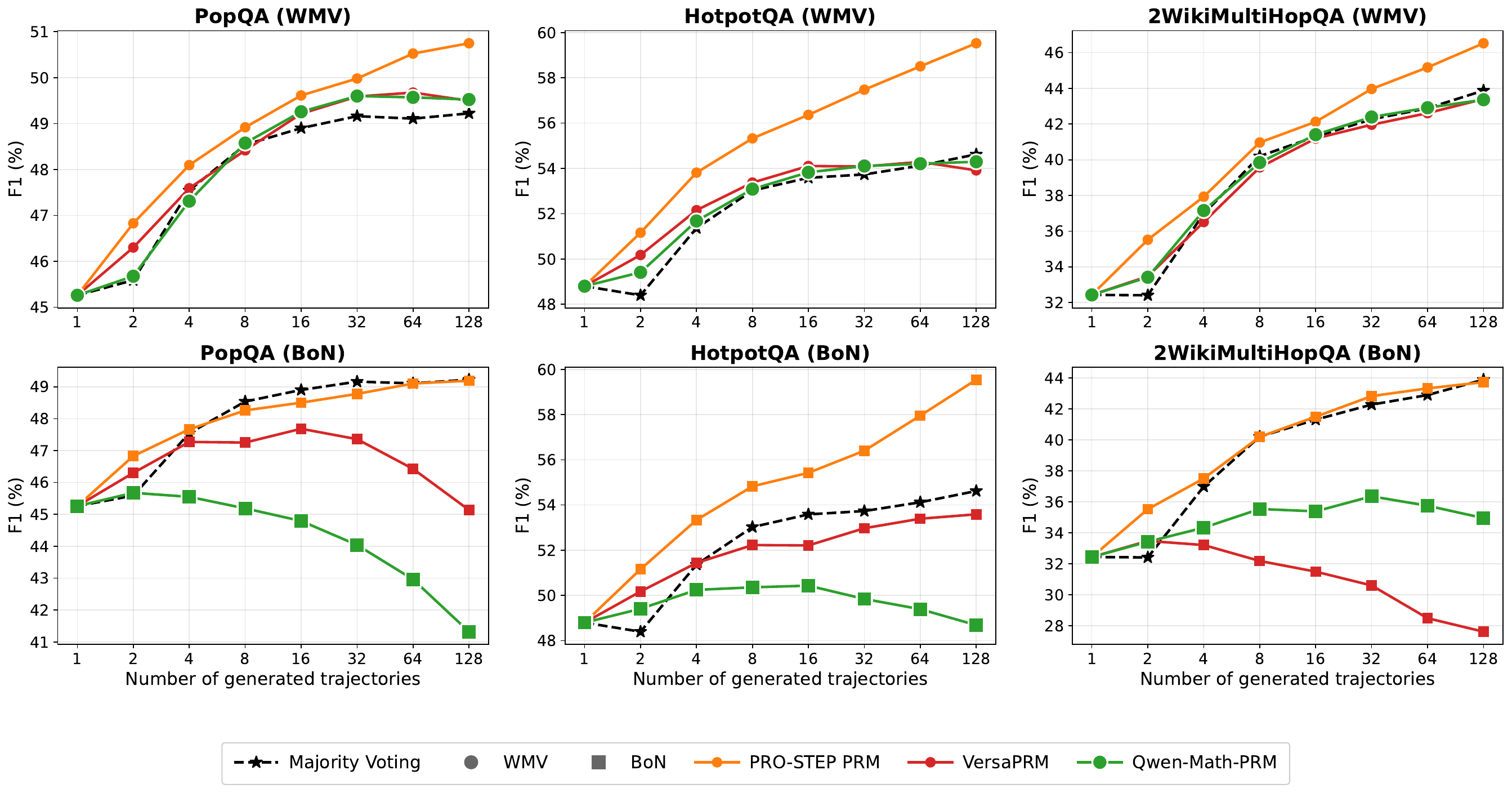}
\caption{Comparison of WMV (top) and BoN (bottom) using a
RAG-specific PRM against open-source PRM approaches across three QA datasets.}
\label{fig:prm_comparison}
\end{figure*}

\paragraph{Process vs. outcome in multi-hop reasoning.} 
Process-level optimization improves performance on datasets requiring extended logical chains. On average, \textsc{PRO-Step} outperforms the outcome-based Search-R1 by 2.5 EM and 3.4 F1 points. This gap widens substantially on complex multi-hop tasks such as 2WikiMultiHopQA, where our method achieves a 9.2 EM and 8.9 F1 improvement. The dense step-level feedback alleviates the reward sparsity problem and improves sample efficiency: \textsc{PRO-Step} surpasses Search-R1 using 5k seed questions, whereas Search-R1 reports roughly 170k training instances.

\paragraph{Comparison with process-reward baselines.}
\textsc{PRO-Step} achieves higher average EM and F1 than ReasonRAG and StepSearch. While these baselines operate at the process level, they optimize each step against the final F1 score, which fails to penalize spurious reasoning where flawed retrieval accidentally produces the correct answer. In contrast, our generative PRM evaluates the logical and evidential validity of each step, filtering out such false positives. Notably, prior methods distill reasoning trajectories from frontier models (e.g., GPT-4o \citep{openai2024gpt4o}), and some additionally invoke multiple LLM calls during inference for document refinement. \textsc{PRO-Step} avoids both, internalizing step-level guidance into the policy through PRM-guided VTS, yielding higher average results with a single model at inference.

\subsection{Effectiveness of Process Reward Model}
\label{sec:component_ablation}
To isolate the contribution of our generative PRM, we compare \textsc{PRO-Step} against a variant without PRM scoring that constructs preference pairs using only the depth-discounted F1 reward. As shown in \Cref{tab:component_ablation}, removing the PRM degrades average performance from 34.5 to 32.7 EM and from 44.1 to 41.9 F1, with the largest drops on multi-hop datasets such as 2WikiMultiHopQA (-2.7 EM) and Bamboogle (-2.4 EM). This larger degradation suggests that PRM supervision is particularly beneficial for multi-hop reasoning. Without PRM evaluation, preference pairs rely solely on final-answer correctness and cannot penalize spurious reasoning that coincidentally leads to correct answers, which the PRM filters out by scoring step-level validity. This validates that considering both logical validity and evidential grounding is essential for RAG. Detailed analysis is provided in \Cref{app:ifbc}.

\subsection{Impact of Optimization Strategies}
\label{sec:ablation}
We compare three optimization strategies on the same process-supervised data to validate that DPO is best suited for process-level learning. \textsc{PRO-Step} (SFT) trains only on the chosen steps $y_t^w$ via next-token prediction. \textsc{PRO-Step} (KTO) uses unpaired binary correctness labels $r_t \in \{0, 1\}$ with Kahneman-Tversky Optimization (KTO) \citep{ethayarajh2024kto}. \textsc{PRO-Step} (DPO) is our default approach using contrastive pairs $y_t^w, y_t^l$.

As shown in \Cref{tab:ablation_results}, DPO achieves the highest average performance with 34.5 EM and 44.1 F1 by providing explicit step-wise contrast between valid and flawed reasoning. SFT lacks contrastive signals and cannot learn to recover from intermediate errors, leading to weaker performance on complex multi-hop tasks such as 11.3 vs. 12.5 EM on MuSiQue. KTO underperforms even SFT on average with 28.0 EM and 36.6 F1, as absolute binary rewards without relative comparison cause instability across datasets such as 20.8 vs. 31.2 EM on Bamboogle. These results confirm that pair-wise contrast, rather than positive-only or unpaired signals, is essential for internalizing the PRM's step-level guidance.

\subsection{Comparison of PRM Approaches on Trajectory Reranking}
\label{sec:prm-reranking}

To validate the necessity of a RAG-specific PRM, we compare our model against existing process reward models. We sample 128 reasoning trajectories per question for 500 instances from PopQA, HotpotQA, and 2WikiMultiHopQA using Qwen2.5-7B-Instruct \citep{qwen2.5}. We evaluate two models, VersaPRM \citep{versaprm2025} and Qwen2.5-Math-PRM-7B \citep{zhang2025lessons}, alongside two trajectory-selection baselines: majority voting (MV), which ignores PRM scores, and weighted majority voting (WMV), which weights each candidate by the minimum step score assigned by a PRM. We also report best-of-$N$ (BoN), which directly selects the highest-scoring trajectory, while scaling the candidate pool size $K$ from 1 to 128. To evaluate the quality of the final selected trajectories, we report token-level F1.

The results in \Cref{fig:prm_comparison} demonstrate that general-purpose and math-specific PRM approaches struggle in RAG environments. Under the BoN strategy, both VersaPRM and MathPRM degrade as $K$ increases, frequently falling below the MV baseline across all three datasets. Because they evaluate logical coherence without verifying factual grounding, these PRM approaches often assign high scores to hallucinated but plausible trajectories. Consequently, as the candidate pool expands, BoN increasingly selects these confidently flawed trajectories.

In contrast, our PRM consistently improves F1 as the candidate pool size increases under WMV. By evaluating both logical validity and the factual grounding of the retrieved documents, our model filters out spurious trajectories. At $K=128$ with WMV, our PRM achieves \textbf{46.5 F1} on 2WikiMultiHopQA, \textbf{59.5 F1} on HotpotQA, and \textbf{50.8 F1} on PopQA, outperforming both MV (43.9 / 54.6 / 49.2) and VersaPRM (43.4 / 53.9 / 49.5) across all three datasets. Under the BoN strategy, our PRM remains stable as $K$ grows, performing comparably to MV on PopQA and 2WikiMultiHopQA (within 0.2 F1) and substantially outperforming it on HotpotQA (+4.9 F1), while VersaPRM and MathPRM degrade sharply at larger $K$. These results validate our motivation for building a RAG-specific PRM that jointly evaluates logical validity and evidential grounding, rather than relying on general-purpose reasoning critics.

\begin{figure}[!t]
    \centering
    \includegraphics[width=\linewidth]{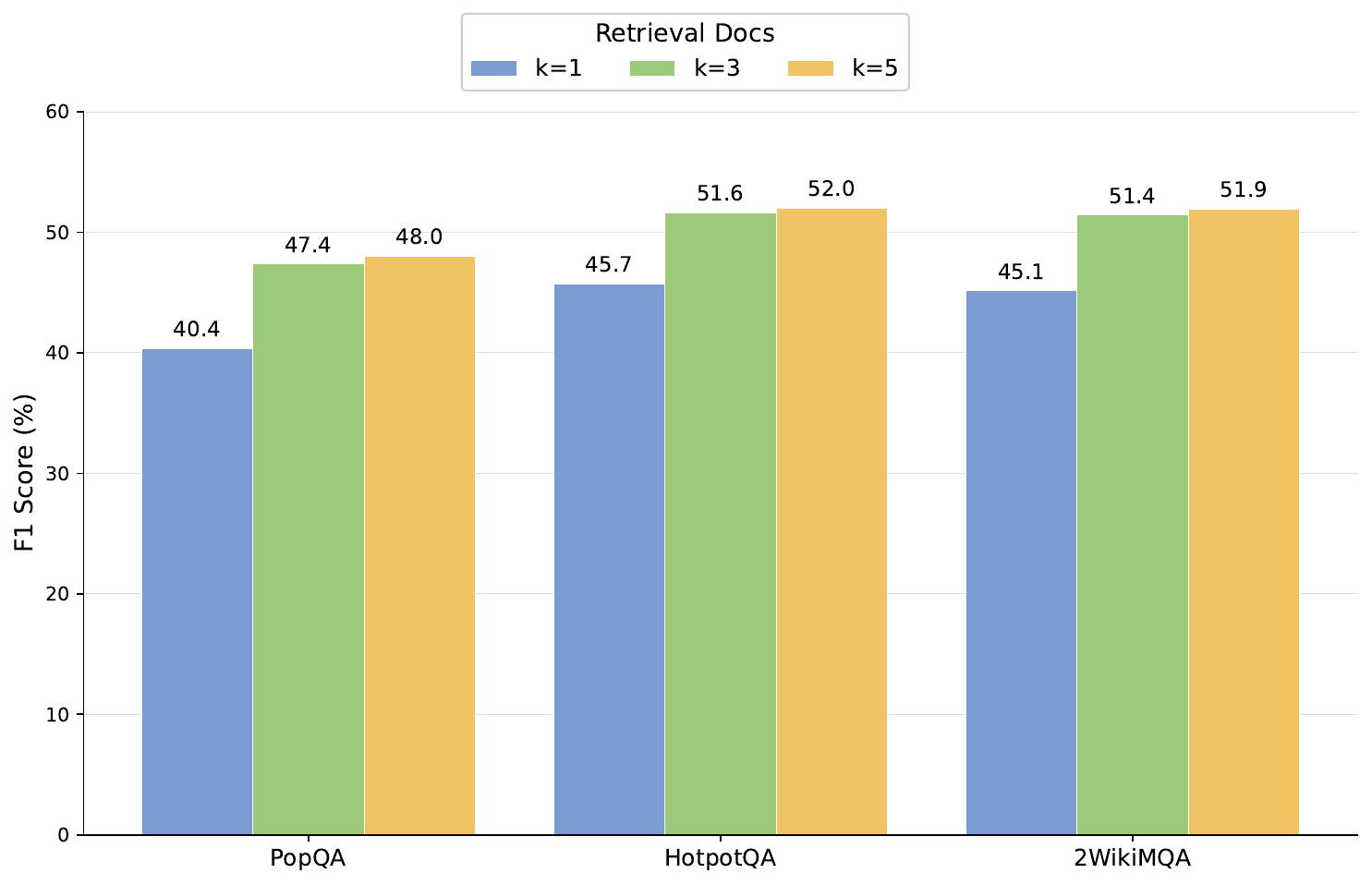}
    \caption{Impact of the number of retrieved documents ($k$) on performance.}
    \label{fig:impact_k}
    
    \vspace{0.5cm} 
    
    \includegraphics[width=\linewidth]{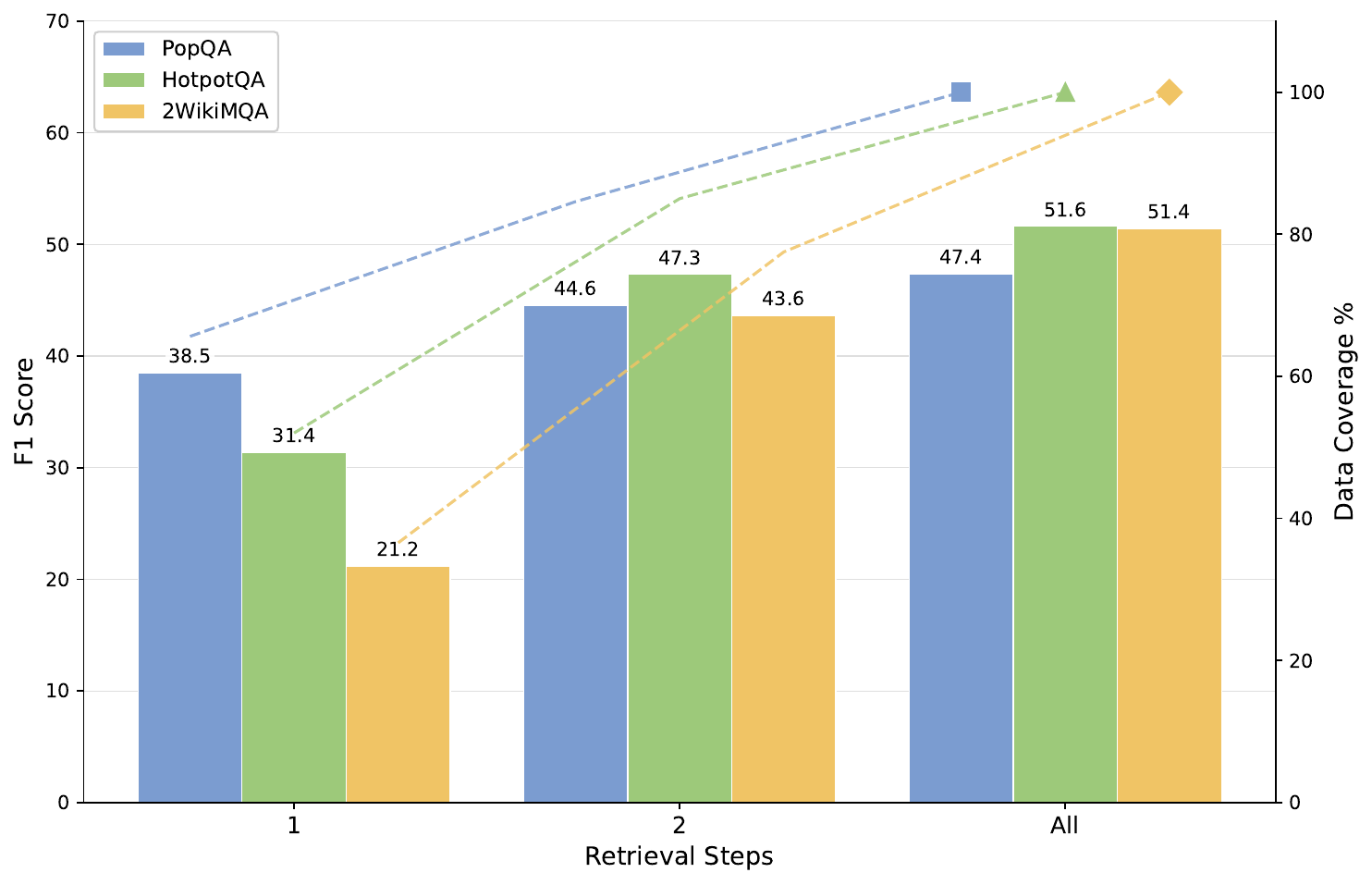}
    \caption{Cumulative average F1 score (bars) and data coverage (dashed lines) across different retrieval steps (1, 2, All) on PopQA, HotpotQA, and 2WikiMultiHopQA.}
    \label{fig:retrieval_steps}
\end{figure}

\subsection{Impact of Retrieval}
\label{sec:impact-retrieval}

In this section, we investigate the performance of \textsc{PRO-Step} under different numbers of retrieved documents and retrieval steps to evaluate its context utilization and planning adaptivity.

\paragraph{Impact of the number of retrieved documents.}
Figure \ref{fig:impact_k} compares \textsc{PRO-Step} under different top-$k$ retrieval settings. Increasing $k$ from 1 to 3 consistently improves performance across all datasets (average +6.4 F1), and expanding to $k=5$ yields further gains. These findings highlight \textsc{PRO-Step}'s capacity to effectively leverage additional retrieved documents, particularly in complex multi-hop scenarios.

\paragraph{Impact of the number of retrieval steps.}
Figure \ref{fig:retrieval_steps} illustrates the cumulative F1 score and data coverage at retrieval-step cap $K$. \textsc{PRO-Step} autonomously adjusts search depth to task difficulty: on single-hop PopQA, 65.6\% of instances are resolved within the first step (cumulative F1 = 38.5), while on multi-hop HotpotQA and 2WikiMultiHopQA only 52.0\% and 36.6\% are resolved at step one, with reasoning naturally extending to additional steps. On 2WikiMultiHopQA, cumulative F1 climbs from 21.2 ($K=1$) to 43.6 ($K=2$) as coverage grows from 36.6\% to 77.5\%, indicating that the model successfully acquires critical missing evidence during its second retrieval. Cumulative F1 increases monotonically across all datasets, demonstrating that \textsc{PRO-Step} mitigates the error-propagation degradation typical of iterative RAG frameworks (qualitative examples in \Cref{app:case-study}).

\subsection{Generalization Across Backbones}
\label{sec:generalization}

Our main experiments use Qwen2.5-7B-Instruct as the policy backbone.

We further examine the generalization of \textsc{PRO-Step} beyond this configuration. Specifically, we evaluate
\textsc{PRO-Step} with a different model family
(Llama-3.1-8B-Instruct), without instruction tuning
(Qwen2.5-7B Base), and at a smaller model scale
(Qwen2.5-3B-Instruct). \Cref{tab:generalization_main} reports results
across all five benchmarks.

\begin{table*}[!t]
\centering
\caption{Generalization of \textsc{PRO-Step} across different policy
backbones. Results are reported using EM and F1.
\textbf{Bold} denotes the better performance within each comparison.}
\label{tab:generalization_main}

\resizebox{\textwidth}{!}{%
\renewcommand{\arraystretch}{1.10}
\begin{tabular}{@{}lcccccccccc|cc@{}}
\toprule
\multirow{2}{*}{\textbf{Method}}
& \multicolumn{2}{c}{\textbf{PopQA}}
& \multicolumn{2}{c}{\textbf{HotpotQA}}
& \multicolumn{2}{c}{\textbf{2WikiMulti}}
& \multicolumn{2}{c}{\textbf{Bamboogle}}
& \multicolumn{2}{c|}{\textbf{MuSiQue}}
& \multicolumn{2}{c}{\textbf{Average}} \\
\cmidrule(lr){2-3}
\cmidrule(lr){4-5}
\cmidrule(lr){6-7}
\cmidrule(lr){8-9}
\cmidrule(lr){10-11}
\cmidrule(lr){12-13}
& EM & F1
& EM & F1
& EM & F1
& EM & F1
& EM & F1
& EM & F1 \\
\midrule

\multicolumn{13}{@{}l}{\textit{Llama-3.1-8B-Instruct}} \\[1pt]

Base
& 5.49 & 16.14
& 5.71 & 14.63
& 5.50 & 12.97
& 7.20 & 15.60
& 1.49 & 6.08
& 5.08 & 13.08 \\

\textsc{PRO-Step}
& \textbf{22.94} & \textbf{28.12}
& \textbf{18.88} & \textbf{27.12}
& \textbf{25.40} & \textbf{31.74}
& \textbf{8.80} & \textbf{16.99}
& \textbf{4.76} & \textbf{8.86}
& \textbf{16.16} & \textbf{22.57} \\

\midrule

\multicolumn{13}{@{}l}{\textit{Qwen2.5-7B Base/Instruct}} \\[1pt]

\textsc{PRO-Step}-Base
& 40.42 & 46.04
& 34.69 & 47.47
& 41.09 & 48.57
& 36.00 & 45.38
& 12.08 & \textbf{22.61}
& 32.85 & 42.01 \\

\textsc{PRO-Step}-Instruct
& \textbf{40.47} & \textbf{47.37}
& \textbf{38.73} & \textbf{51.63}
& \textbf{44.07} & \textbf{51.43}
& \textbf{36.80} & \textbf{47.63}
& \textbf{12.49} & 22.41
& \textbf{34.51} & \textbf{44.09} \\

\midrule

\multicolumn{13}{@{}l}{\textit{Qwen2.5-3B-Instruct}} \\[1pt]

Search-R1
& 36.87 & 42.14
& 31.82 & 41.30
& 35.96 & 41.86
& 18.40 & 27.97
& 8.07 & 14.27
& 26.22 & 33.51 \\

\textsc{PRO-Step}
& \textbf{39.88} & \textbf{46.10}
& \textbf{32.82} & \textbf{44.33}
& \textbf{44.10} & \textbf{50.77}
& \textbf{28.00} & \textbf{38.96}
& \textbf{10.38} & \textbf{19.85}
& \textbf{31.04} & \textbf{40.00} \\

\bottomrule
\end{tabular}%
}
\end{table*}

\begin{table}[t]
    \centering
    \caption{Final EM on recovered trajectories. $n$ is given as
    Search-R1\,/\,\textsc{PRO-Step}.}
    \label{tab:recovery}
    \small
    \renewcommand{\arraystretch}{1.10}
    \begin{tabular*}{\columnwidth}{@{\extracolsep{\fill}} lccc @{}}
        \toprule
        \textbf{Dataset} & \textbf{$n$} & \textbf{Search-R1} & \textbf{\textsc{PRO-Step}} \\
        \midrule
        HotpotQA   & 1{,}726\,/\,1{,}081 & 51.3 & \textbf{54.4} \\
        PopQA      & 915\,/\,623         & 44.6 & \textbf{49.6} \\
        2WikiMulti & 4{,}002\,/\,2{,}945 & 47.4 & \textbf{52.8} \\
        Bamboogle  & 34\,/\,25           & 82.4 & \textbf{84.0} \\
        MuSiQue    & 550\,/\,418         & 38.4 & \textbf{38.5} \\
        \bottomrule
    \end{tabular*}
\end{table}

As shown in \Cref{tab:generalization_main}, \textsc{PRO-Step}
improves Llama-3.1-8B-Instruct across all five datasets, increasing
the average performance from 5.08 / 13.08 to 16.16 / 22.57 EM / F1
(+11.1 EM and +9.5 F1). This shows that the gains are not restricted
to the Qwen model family.

Without instruction tuning, Qwen2.5-7B Base with \textsc{PRO-Step}
achieves 32.85 / 42.01 average EM / F1, compared with
34.51 / 44.09 for the main instruction-tuned backbone. The relatively
small average difference (1.66 EM and 2.08 F1) shows that the framework
retains most of its performance without an instruction-tuned
initialization.

At the 3B scale, \textsc{PRO-Step} outperforms Search-R1 across all
five datasets, improving the average from 26.22 / 33.51 to
31.04 / 40.00 EM / F1 (+4.82 EM and +6.49 F1). Together, these
results show that \textsc{PRO-Step} extends across different model
families, alignment settings, and model scales. 
 The 3B configuration additionally uses an SFT warmup on trajectories from the 7B policy. 
Additional 3B results and training analysis are provided in \Cref{app:backbone}.

\subsection{Recovery from Retrieval Failure}
\label{sec:recovery}
Outcome-based optimization lacks a direct signal to detect or penalize
intermediate retrieval failures, allowing early errors to propagate
through subsequent reasoning steps. To examine this failure mode, we
evaluate \textsc{PRO-Step} and the outcome-based Search-R1 on
\emph{recovered trajectories}, where the initial retrieval fails to
surface the gold passage but a later retrieval succeeds. Since this
subset is determined by each model's own search behavior, the two
systems are evaluated on overlapping but non-identical question sets.

As shown in \Cref{tab:recovery}, \textsc{PRO-Step} achieves higher
final EM than Search-R1 on all five datasets, and its subsets are
consistently smaller because it fails less often at the initial
retrieval. On 2WikiMulti, the gains hold at every recovery depth
$k$ with sufficient support (+4.1, +8.3, and +6.8 EM at $k=1,2,3$),
and \textsc{PRO-Step} concentrates its recoveries earlier, resolving
87\% of its subset at $k=1$ versus 80\% for Search-R1. The advantage is not driven by subset selection, as \textsc{PRO-Step} also outperforms Search-R1 when the initial retrieval succeeds, by 12.1 EM on 2WikiMulti and 29.2 EM on Bamboogle.
Detailed results are provided in \Cref{app:recovery}. These results show that
\textsc{PRO-Step} better recovers from retrieval failures, mitigating
error propagation through subsequent reasoning steps.
\section{Conclusion}
\label{sec:conclusion}

We introduced \textsc{PRO-Step}, a framework that integrates process-level supervision into RAG to mitigate error propagation in multi-hop reasoning. Our approach trains a generative PRM to evaluate both the logical validity and the evidential grounding of intermediate steps. We employ this PRM to guide VTS, proactively constructing step-level preference pairs by explicitly filtering out spurious trajectories that coincidentally reach correct final outcomes. Using these targeted preference pairs, we optimize the policy model using DPO to enhance its ability to navigate interleaved generation and retrieval. Experiments demonstrate that \textsc{PRO-Step} addresses intermediate reasoning flaws and achieves the best average EM and F1 across the five evaluated QA benchmarks. By mitigating the limitations of sparse outcome rewards, \textsc{PRO-Step} offers an effective approach for fine-grained policy optimization in multi-hop QA tasks.

\section*{Limitations}

\paragraph{Limitations of QwQ-32B as a PRM labeler.}
While frontier closed-source models (e.g., GPT-4o) might marginally reduce potential label noise, we utilize QwQ-32B to ensure our pipeline is fully reproducible without restrictive API licenses. To validate this choice and quantify reliability, we conducted a 500-sample audit against Claude Opus 4.7 (detailed in \Cref{app:prm_validation}). The results demonstrated substantial agreement ($\kappa = 0.6104$) without systematic permissive bias, confirming that our open-source approach democratizes process-supervised training without sacrificing critical step-level accuracy.

\paragraph{Trade-offs of open-source, small-data supervision.}
While \textsc{PRO-Step} achieves the highest average EM and F1, it does not uniformly dominate every dataset. The largest gap appears on MuSiQue, where \textsc{PRO-Step} trails StepSearch by 1.3 EM. We attribute this to our deliberate choice to avoid closed-source teachers and train on a smaller pool (5{,}000 seed questions, 15{,}877 pairs) than StepSearch's GPT-4o-distilled 19k corpus. This is a transparent trade-off: \textsc{PRO-Step} is reproducible without API access, at the cost of some headroom on the hardest tasks.

\paragraph{Bamboogle's small test size limits statistical reliability.}
Bamboogle contains only 125 test instances, yielding wide confidence intervals. Our paired $t$-test against ReasonRAG returns $p = 0.66$ (\Cref{app:significance}), indicating the apparent gap is statistically indistinguishable from noise.

\paragraph{PRM-feedback-guided regeneration disrupts policy alignment.}
We explored augmenting preference pairs via PRM-feedback-guided regeneration \citep{genprm2026}, but the augmented variant degrades evaluation by 10.7 EM and 9.8 F1 on average (\Cref{app:regen}). Regenerated trajectories drift from the natural search-and-reason pattern of inference-time generation, and the outcome-margin filter cannot detect this stylistic drift. This motivates future work on action-format-consistent augmentation.

\section*{Ethical Considerations}
\label{sec:ethics}
This work presents \textsc{PRO-Step}, a framework for optimizing retrieval-augmented 
reasoning in large language models. All datasets used in this study 
(HotpotQA, MuSiQue, 2WikiMultiHopQA, PopQA, Bamboogle) are publicly available 
benchmarks intended for research purposes. Our training data is derived solely 
from these existing datasets, and no personally identifiable information is 
collected or processed. The proposed framework does not introduce new risks beyond 
those inherent to LLM-based systems, such as residual hallucination or retrieval 
of outdated information. Practitioners deploying this framework in 
domains should apply appropriate human oversight.

AI writing assistance was used solely for grammar 
and language checking of the manuscript.

\section*{Acknowledgments}

This research was supported by the Basic Science Research Program through the
National Research Foundation of Korea (NRF) funded by the Korean government
(MSIT) (No. IITP-2026-RS-2024-00346737), and by grants from the Institute for
Information \& Communications Technology Planning \& Evaluation (IITP) funded
by the Ministry of Science and ICT (MSIT), Korea, through the Global Scholars
Invitation Program (No. IITP-2026-RS-2024-00459638), the Graduate School of
Metaverse Convergence at Sungkyunkwan University
(No. IITP-2026-RS-2023-00254129), and the ICT Challenge and Advanced Network
of HRD (ICAN) support program (No. IITP-2026-RS-2023-00259497).

\bibliography{custom}
\clearpage 
\appendix
\appendix

\section*{Appendix Overview}

The appendix includes the following sections:

\begin{itemize}
\item \textbf{\Cref{app:Prompt}: Prompts and Models Reference.} Details the prompts used for the policy model's trajectory generation, the PRM's training-data annotation, and the PRM's VTS inline scoring. This serves as a supplement to \Cref{sec:method}.

\item \textbf{\Cref{app:data-stats}: Data Statistics.} Provides detailed statistics of the PRM training corpus and the DPO preference pairs, along with an ablation on the outcome-margin filter. This serves as a supplement to \Cref{sec:method,sec:experiments}.

\item \textbf{\Cref{app:VTS}: PRM-guided Value Tree Search.}
Presents the mathematical formulations and hyperparameters for the
PRM-guided value tree search, together with a branching-factor ablation.
This serves as an extended explanation of \Cref{sec:policy-opt}.

\item \textbf{\Cref{app:implementation}: Implementation Details and Hyperparameters.} Reports the full PRM weight ($\alpha$) sweep and training hyperparameters for SFT, DPO, and KTO. This serves as a supplement to \Cref{sec:experiments,sec:ablation}.

\item \textbf{\Cref{app:prm_validation}: Direct Validation of the PRM.}
We complement the indirect PRM evaluation in
\Cref{sec:prm-reranking} with direct validation of label quality using
both human and frontier-model audits.

\item \textbf{\Cref{app:significance}: Statistical Significance of Main Results.} Provides bootstrap confidence intervals and per-dataset paired $t$-tests against prior baselines. This serves as a supplement to \Cref{tab:main_results}.

\item \textbf{Section~\ref{app:backbone}: Generalization Across Model
Families, Scales, and Alignments.} Provides additional 3B analysis,
including base-model results, statistical significance, and training
details, supplementing Section~\ref{sec:generalization}.

\item \textbf{\Cref{app:ifbc}: Intermediate-Flawed-But-Correct Analysis.}
Quantifies trajectories that contain flawed intermediate steps despite
reaching the correct final answer, directly evaluating the failure mode
targeted by process-level supervision.

\item \textbf{Section~\ref{app:recovery}: Detailed Recovery Analysis.}
Provides the full breakdowns of the recovery experiment, including
token-level F1, per-depth results, and the complementary case where
the initial retrieval already succeeds. This serves as a supplement to
Section~\ref{sec:recovery}.

\item \textbf{\Cref{app:case-study}: Case Study.} Presents two qualitative examples illustrating how PRM step-level supervision prevents entity confusion and error propagation. This serves as an extended example of \Cref{sec:experiments}.

\item \textbf{\Cref{app:regen}: PRM-Feedback-Guided Regeneration.} Documents a failed augmentation strategy in which PRM rationales were used to regenerate training trajectories, along with a diagnosis of the resulting distributional mismatch. This serves as an extended discussion of the limitations.

\item \textbf{\Cref{app:resources}: Released Resources.}
Provides links to the released implementation, models, and training datasets. 

\end{itemize}

\section{Prompts and Models Reference}
\label{app:Prompt}

We provide the prompts used for policy trajectory generation,
PRM training-data annotation, and PRM-guided VTS evaluation.


\begin{figure*}[!t]
\centering
\begin{promptbox}{policyblue}
{Policy Model: Trajectory Generation}

\textbf{System Prompt.}
You are a helpful assistant who is good at answering questions with
multi-turn search engine calling. To answer questions, you must first
reason through the available information using
\texttt{<think>} and \texttt{</think>}. If you identify missing
knowledge, you may issue a search request using
\texttt{<search> query </search>} at any time. The retrieval system
will provide you with relevant documents enclosed in
\texttt{<documents>} and \texttt{</documents>}. You can search as many
times as you want. Once you have sufficient information or if you find
no further external knowledge is needed, directly provide your final
answer. Ensure your answer is concise, using nouns or short phrases
whenever possible. Conclude with:
``So the answer is \texttt{<answer>answer</answer>}''.

\tcblower

\textbf{User Prompt.}

\texttt{Question: \{question\}}

\end{promptbox}
\end{figure*}


\begin{figure*}[!t]
\centering
\begin{promptbox}{labelgreen}
{PRM Training-Data Generation: QwQ-32B Annotation Prompt}
\label{app:labeling_rubric}

This prompt is used by QwQ-32B for full-trajectory evaluation and
generation of PRM training labels. The gold answer is provided as a
reference, while intermediate steps are evaluated according to the
criteria below.

\medskip
\textbf{System Prompt.}

You are a strict process supervisor for a multi-hop question answering
agent that uses retrieval-augmented generation (RAG). Your task is to
evaluate each step of the agent's trajectory and assign a binary label:
GOOD or BAD.

The agent operates using four XML-tagged actions:
\texttt{<think>} (Internal reasoning),
\texttt{<search>} (Retrieval query),
\texttt{<documents>} (Retrieved passages), and
\texttt{<answer>} (Final answer).

\medskip
\textbf{Labeling Principles}

\begin{enumerate}
    \item Default label is BAD. Assign GOOD only when a step makes a
    verifiable contribution toward answering.

    \item No information gain means BAD (e.g., restating the question,
    repeating prior reasoning).

    \item Do not use the model's own world knowledge. Unverified claims
    not in \texttt{<documents>} are hallucination.
\end{enumerate}

\textbf{Evaluation Criteria}

\begin{description}
    \item[\textbf{R1. Entity/Relation.}]
    BAD if \texttt{<think>} misidentifies an entity, reverses a
    relation, or drifts. GOOD if it maintains correct grounding.

    \item[\textbf{R2. Search.}]
    GOOD if the query is specific, returns relevant
    \texttt{<documents>}, or tries a new strategy after a failure.
    BAD if it repeats a failed attempt, is vague, or returns irrelevant
    data.

    \item[\textbf{R3. Reasoning.}]
    GOOD if it extracts useful information from
    \texttt{<documents>} or identifies a concrete knowledge gap.
    BAD if it merely restates, repeats, or hallucinates.

    \item[\textbf{R4. Answer.}]
    BAD if empty, uncertain, requires unsupported logical leaps, or
    contradicts \texttt{<documents>}. GOOD if specific and logically
    derivable.

    \item[\textbf{R5. Recovery.}]
    GOOD if the agent issues a new, meaningfully different
    \texttt{<search>} after a failure. BAD if it proceeds to
    \texttt{<answer>} while ignoring the failure.

    \item[\textbf{R6. Overconfidence.}]
    BAD if the agent produces \texttt{<answer>} without prior
    successful search.
\end{description}

\tcblower

\textbf{User Prompt.}

\texttt{Question: \{question\}} \qquad
\texttt{Ground Truth: \{gold\_answer\}}

\medskip
\textbf{\#\# Trajectory}

\texttt{\{full trajectory with all steps\}}

\medskip
\textbf{Expected Output.}

Return a JSON array:

\texttt{[\{"step": 1, "label": "GOOD", "reasoning": "..."\}, ...]}

\end{promptbox}
\end{figure*}


\begin{figure*}[!t]
\centering
\begin{promptbox}{prmorange}
{PRM-Guided VTS: Inline Step Evaluation Prompt}

This prompt is used by the deployed 8B PRM for step-level evaluation
during VTS expansion.

\medskip
\textbf{System Prompt.}

You are a step-level PRM for evaluating reasoning quality in multi-hop
question answering. The trajectory uses XML tags:
\texttt{<think>}, \texttt{<search>}, \texttt{<answer>}, and
\texttt{<documents>}. Analyze each step's logical soundness and
evidence grounding. First explain your reasoning inside
\texttt{[REASONING]} tags, then output a label
(1=good, 0=bad).

\tcblower

\textbf{User Prompt.}

\texttt{Question: \{question\}}

\medskip
\textbf{Previous Steps}

\textbf{\#\# Step 1}

\texttt{<think>...</think>} \qquad
\texttt{<search>...</search>} \qquad
\texttt{<documents>...</documents>}

\[
\cdots
\]

\textbf{Current Step to Evaluate}

\textbf{\#\# Step N}

\texttt{<think>...</think>} \qquad
\texttt{<search>...</search>}

\medskip
\textbf{Expected Output Format}

\texttt{[REASONING]} \{PRM reasoning\}
\texttt{[/REASONING]}

\texttt{Label: \{1 or 0\}}

\end{promptbox}
\end{figure*}
\FloatBarrier

\section{Data Statistics}
\label{app:data-stats}

We present detailed statistics for the datasets used in the two
training stages. To avoid ambiguity, we distinguish between the
original seed-question pool, the subset retained after filtering, and
the final number of step-level supervision instances.

\paragraph{PRM Training Data.}
The PRM is trained on 109{,}664 step-level decisions extracted from
31{,}728 reasoning trajectories sampled across 2{,}000 source
multi-hop QA questions. \Cref{tab:prm_stats} summarizes the
action-type distribution and the QwQ-32B GOOD/BAD verdicts. The
majority of the steps are search actions (71.2\%), reflecting the
highly interactive nature of our multi-hop RAG environment. The
\texttt{Reason} action (internal thought) accounts for only $0.5\%$
of the steps and has a high rejection rate (73.7\% BAD), indicating
that the PRM penalizes unsupported internal hallucinations and
forces the model to rely on external search. We validate the
reliability of these QwQ-generated annotations through an
independent frontier-LLM audit reported in
\Cref{app:prm_validation}.

\begin{table}[ht]
\centering
\small
\resizebox{\columnwidth}{!}{%
\begin{tabular}{@{}lrcrcr@{}}
\toprule
\textbf{Step Type} & \multicolumn{2}{c}{\textbf{Count (\%)}} & \multicolumn{1}{c}{\textbf{GOOD}} & \multicolumn{2}{c}{\textbf{BAD}} \\
\midrule
Search & 78{,}055 & (71.2\%) & 55{,}710 (71.4\%) & 22{,}345 & (28.6\%) \\
Answer & 31{,}062 & (28.3\%) & 14{,}436 (46.5\%) & 16{,}626 & (53.5\%) \\
Reason &     547  &  (0.5\%) &     144 (26.3\%) &     403 & (73.7\%) \\
\midrule
\textbf{Total} & \textbf{109{,}664} & \textbf{(100\%)} & \textbf{70{,}290 (64.1\%)} & \textbf{39{,}374} & \textbf{(35.9\%)} \\
\bottomrule
\end{tabular}%
}
\caption{Statistics of the PRM training corpus: 109{,}664 steps
generated from 31{,}728 trajectories over 2{,}000 source questions
(averaging 3.5 steps per trajectory). Labels assigned by QwQ-32B
under the R1--R6 rubric (\Cref{app:labeling_rubric}).}
\label{tab:prm_stats}
\end{table}

\paragraph{DPO Preference Pairs.}
For policy optimization, we began from the full pool of 5{,}000
seed multi-hop QA questions described in \Cref{sec:experiments}.
After running VTS, we retained only preference pairs satisfying an outcome-margin filter (chosen $F_1 \geq 0.2$ and $\Delta F_1 \geq 0.2$), yielding 2{,}866 unique questions and 15{,}877 contrastive pairs, or approximately $5.5$ pairs per retained question.

We adopt this filter for training and data efficiency rather than for performance. As shown in \Cref{tab:margin_ablation}, removing the filter yields 39{,}578 pairs ($2.5\times$ more data) with comparable average performance (34.65 vs.\ 34.51 EM), indicating that the discarded low-margin pairs contribute little additional signal relative to their training cost.

As detailed in \Cref{tab:dpo_stats}, the DPO pairs exhibit
rich diversity in action types, tree depths, and reward margins. About
three-quarters ($75.0\%$) of the comparisons pit actions of the same
type against each other (Search vs.\ Search: $57.5\%$, Answer vs.\
Answer: $17.5\%$), allowing the policy to learn fine-grained strategic
differences rather than mere format adherence. The step-level
distribution peaks at depth 2 ($45.7\%$) and depth 3 ($29.5\%$),
directly aligning with the multi-hop reasoning requirements. The
reward difference between chosen and rejected steps ensures a robust
gradient: $98.4\%$ of pairs have a substantial margin
($\geq 0.20$), with over half ($52.0\%$) exhibiting a high margin
($\geq 0.50$).

\begin{table*}[!ht]
\centering
\caption{Effect of the outcome-margin filter on data scale and downstream performance. Removing the filter increases the pair count by $2.5\times$ but yields comparable average performance.}
\label{tab:margin_ablation}
\resizebox{\textwidth}{!}{%
\renewcommand{\arraystretch}{1.10}
\begin{tabular}{@{}lcccccccc@{}}
\toprule
\textbf{Variant} & \textbf{Pairs} & \textbf{HotpotQA} & \textbf{PopQA} & \textbf{2WikiMulti} & \textbf{Bamboogle} & \textbf{MuSiQue} & \textbf{Average} \\
\midrule
\textsc{PRO-Step} (with margin filter) & 15{,}877 & 38.73 / 51.63 & 40.47 / 47.37 & 44.07 / 51.43 & 36.80 / 47.63 & 12.49 / 22.41 & \textbf{34.51 / 44.09} \\
w/o margin filter & 39{,}578 & 37.62 / 50.58 & 39.20 / 46.40 & 44.47 / 52.62 & 39.20 / 47.70 & 12.78 / 23.16 & 34.65 / 44.09 \\
\bottomrule
\end{tabular}%
}
\end{table*}

\begin{table}[ht]
\centering
\small
\resizebox{\columnwidth}{!}{%
\begin{tabular}{@{}lrlr@{}}
\toprule
\textbf{Pair Type}          & \textbf{\%} & \textbf{Reward Diff Margin} & \textbf{\%} \\
\midrule
Search vs.\ Search          & 57.5        & $<0.05$ (Low)               & 0.3         \\
Answer vs.\ Answer          & 17.5        & $0.05$--$0.10$              & 0.6         \\
Answer vs.\ Search          & 16.5        & $0.10$--$0.20$              & 0.6         \\
Search vs.\ Answer          & 4.1         & $0.20$--$0.50$              & 46.4        \\
Other                       & 4.4         & $\geq 0.50$ (High)          & 52.0        \\
\midrule
\textbf{Step Level (Depth)} & \textbf{\%} & \textbf{Action (Ch / Rej)}  & \textbf{\%} \\
\midrule
Step 1                      & 24.0        & Search                      & 61.9 / 74.1 \\
Step 2                      & 45.7        & Answer                      & 37.7 / 21.7 \\
Step 3                      & 29.5        & Reason                      & 0.4 / 4.2   \\
Step 4                      & 0.8         & --                          & --          \\
\bottomrule
\end{tabular}%
}
\caption{Detailed statistics of the 15{,}877 DPO preference pairs
under the outcome filter (chosen $F_1 \geq 0.2$ AND
$\Delta F_1 \geq 0.2$). Reward margin is computed on
$V(s) = \bar{Q}(s) + \alpha \cdot \hat{r}(s)$ with $\alpha = 0.3$.
Action percentages represent the proportion of Chosen / Rejected
actions respectively.}
\label{tab:dpo_stats}
\end{table}

\section{VTS Implementation Details}
\label{app:VTS}

This section provides the detailed mathematical formulations and
algorithmic mechanics for the VTS procedure introduced in
\Cref{sec:policy-opt}. Specifically, we detail how the step-level
feedback from our generative PRM is integrated into the tree search to
actively balance exploration and exploitation, ultimately guiding the
generation of high-quality, process-supervised trajectories for
\textsc{PRO-Step}.

\paragraph{Selection.}
\begin{equation*}
\mathrm{UCB}(s_t) = \bar{Q}(s_t)
  + c\sqrt{\frac{\ln N(\mathrm{parent}(s_t))}{N(s_t)}}
\end{equation*}
where $N(s_t)$ is the visit count of $s_t$,
$\bar{Q}(s_t) = W(s_t)/N(s_t)$ is its mean backed-up value, $W(s_t)$
is its cumulative backed-up value, and $c$ is the exploration
constant. Unvisited nodes ($N=0$) are selected immediately.

\paragraph{Terminal reward.}
\begin{equation*}
Q(s_T) = \mathrm{F1}(s_T,\, y^*) \cdot \gamma^{d(s_T)}
\end{equation*}
where $\gamma \in (0,1)$ discourages unnecessarily long trajectories
and $d(\cdot)$ denotes node depth.

\paragraph{Back-propagation.}
After evaluating a terminal node, we propagate $Q(s_T)$ to all
ancestors:
\begin{equation*}
N(s) \leftarrow N(s) + 1, \qquad W(s) \leftarrow W(s) + Q(s_T)
\end{equation*}
When $\pi_\psi$ evaluates a non-terminal node $s_t$, we apply a
retroactive update to its ancestors without changing their visit
counts:
\begin{equation*}
W(s) \leftarrow W(s) + \hat{r}(s_t) \cdot \gamma^{d(s_t)}
\end{equation*}

\paragraph{VTS Parameters.}
During the process-supervised data generation phase, we utilize VTS
to explore reasoning trajectories. \Cref{tab:VTS_params} details the
specific hyperparameter values used for the tree search. To balance
the reward scale, the terminal F1 score is discounted by $0.9$ per
depth level, and the intermediate step score is computed as a weighted
sum of the backed-up F1 reward and the PRM's feedback ($\alpha = 0.3$).
Because the PRM provides binary labels $\{0, 1\}$ whereas the F1 score
is continuous in $[0, 1]$, this weight allows the process reward to
break ties among steps with similar outcomes without overriding
substantial F1 differences.

\begin{table}[ht]
\centering
\small
\resizebox{\columnwidth}{!}{%
\begin{tabular}{@{}ll@{}}
\toprule
\textbf{Parameter} & \textbf{Value} \\
\midrule
Branching factor ($K$) & 3 \\
Max depth & 7 \\
Iterations per question & 64 \\
F1 discount ($\gamma$) & 0.9 (Reward $= \mathrm{F1} \cdot 0.9^{d}$) \\
PRM weight ($\alpha$) & 0.3 (Combined $= \mathrm{Reward} + 0.3 \cdot \mathrm{PRM}$) \\
PRM scoring interval & Every 4 iterations \\
Retrieved docs per search & 3 \\
Preference margin ($\delta$) & 0.01 \\
\bottomrule
\end{tabular}%
}
\caption{Hyperparameters for the VTS rollout phase.}
\label{tab:VTS_params}
\end{table}
\paragraph{Branching-factor ablation.}
We additionally evaluate the effect of the branching configuration used
during VTS trajectory generation. As shown in
\Cref{tab:branching_ablation}, the main configuration improves average
performance by 1.9 EM and 2.0 F1 over the reduced-branching setting.
The reduced setting nevertheless remains competitive, indicating that
the effectiveness of \textsc{PRO-Step} is not tied to a single
branching configuration.

\begin{table*}[!ht]
\centering
\caption{Ablation of the VTS branching configuration. The main
configuration improves average performance by 1.9 EM and 2.0 F1 over
the reduced-branching setting.}
\label{tab:branching_ablation}
\small
\resizebox{\textwidth}{!}{%
\begin{tabular}{@{}lcccccc@{}}
\toprule
\textbf{Configuration}
& \textbf{PopQA}
& \textbf{HotpotQA}
& \textbf{2WikiMulti}
& \textbf{Bamboogle}
& \textbf{MuSiQue}
& \textbf{Average} \\
\midrule
Reduced branching ($K=2$)
& 39.5 / 46.5
& 38.0 / 50.3
& 39.6 / 48.4
& 32.8 / 42.5
& \textbf{13.2} / \textbf{23.0}
& 32.6 / 42.1 \\

Main configuration ($K=3$)
& \textbf{40.5} / \textbf{47.4}
& \textbf{38.7} / \textbf{51.6}
& \textbf{44.1} / \textbf{51.4}
& \textbf{36.8} / \textbf{47.6}
& 12.5 / 22.4
& \textbf{34.5} / \textbf{44.1} \\
\bottomrule
\end{tabular}}
\end{table*}
\section{Implementation Details and Hyperparameters}
\label{app:implementation}

\paragraph{PRM Weight Sweep.}
\Cref{tab:alpha_sweep} reports the full sweep over $\alpha$ for the combined value $V(s) = \bar{Q}(s) + \alpha \cdot \hat{r}(s)$. The results justify our use of PRM-based step evaluation: every configuration that incorporates PRM signals ($\alpha > 0$, except $\alpha = 0.5$) outperforms the no-PRM baseline ($\alpha = 0$, 32.68 EM), with $\alpha = 0.3$ achieving the best average (34.51 EM, 44.09 F1). This confirms that PRM signals provide supervision beyond what outcome rewards alone can offer, and that process and outcome rewards function as complementary signals.

\begin{table*}[!t]
\centering
\caption{Performance comparison across different Process Reward Model
(PRM) weight ($\alpha$) values. The table reports EM / F1 scores
across 5 benchmarks. $\star$ denotes the selected configuration.}
\label{tab:alpha_sweep}
\resizebox{\textwidth}{!}{%
\renewcommand{\arraystretch}{1.10}
\begin{tabular}{@{}lcccccc@{}}
\toprule
\textbf{$\alpha$} & \textbf{HotpotQA} & \textbf{PopQA} & \textbf{2WikiMultiHopQA} & \textbf{Bamboogle} & \textbf{MuSiQue} & \textbf{Average} \\
\midrule
0.0 (no PRM) & 36.75 / 49.31 & 38.48 / 45.01 & 41.44 / 48.40 & \underline{34.40} / \underline{44.20} & 12.33 / 22.35 & 32.68 / 41.85 \\
\textbf{0.3 $\star$} & \textbf{38.73} / \textbf{51.63} & \textbf{40.47} / \textbf{47.37} & \textbf{44.07} / \textbf{51.43} & \textbf{36.80} / \textbf{47.63} & \underline{12.49} / \underline{22.41} & \textbf{34.51} / \textbf{44.09} \\
0.5 & 34.95 / 47.53 & 36.62 / 44.33 & 41.28 / 48.21 & 33.60 / 41.91 & 10.47 / 20.41 & 31.38 / 40.48 \\
1.0 & \underline{37.97} / \underline{50.60} & \underline{39.94} / \underline{46.10} & \underline{42.68} / \underline{50.38} & 32.80 / 41.39 & \textbf{13.49} / \textbf{23.46} & \underline{33.38} / \underline{42.39} \\
\bottomrule
\end{tabular}%
}
\end{table*}

\paragraph{Training Hyperparameters.}
\Cref{tab:training_params} summarizes the hyperparameter
configurations for training the policy model across different
optimization strategies (SFT, DPO, KTO). All policy models share the
same base model (Qwen2.5-7B-Instruct) and apply document masking
during training to prevent the model from memorizing the retrieved
context. We utilize LoRA for parameter-efficient fine-tuning across
all models, targeting all linear layers.

\begin{table}[ht]
\centering
\small
\resizebox{\columnwidth}{!}{%
\begin{tabular}{@{}lccc@{}}
\toprule
\textbf{Parameter}            & \textbf{SFT}      & \textbf{DPO}            & \textbf{KTO} \\
\midrule
\textbf{Base Model}           & \multicolumn{3}{c}{Qwen2.5-7B-Instruct} \\
\textbf{Data Size}            & 15{,}877 (Chosen) & 15{,}877 pairs          & 31{,}754 (15{,}877 +/15{,}877 -) \\
\textbf{Objective}            & Next-token        & Sigmoid ($\beta=0.1$)   & KTO ($\beta=0.1$) \\
\textbf{Learning Rate}        & 2e-5              & 2e-5                    & 5e-6 \\
\textbf{Effective Batch Size} & 16                & 16                      & 16 \\
\textbf{Epochs}               & 1                 & 1                       & 1 \\
\textbf{Max Length}           & 8{,}192           & 8{,}192                 & 8{,}192 \\
\textbf{Document Masking}     & Yes               & Yes                     & Yes \\
\midrule
\multicolumn{4}{c}{\textbf{LoRA Configuration}} \\
\midrule
\textbf{Rank ($r$)}           & 64                & 64                      & 64 \\
\textbf{Alpha ($\alpha$)}     & 128               & 128                     & 128 \\
\textbf{Dropout}              & 0.05              & 0.05                    & 0.05 \\
\textbf{Target Modules}       & \multicolumn{3}{c}{all-linear} \\
\bottomrule
\end{tabular}%
}
\caption{Training hyperparameters for the policy models. The effective
batch size of 16 is achieved via $1 \times 8$ gradient accumulation
$\times$ 2 GPUs across all three objectives. KTO uses both chosen and
rejected from the outcome filter pair set as desirable / undesirable samples
(hence 31{,}754 total).}
\label{tab:training_params}
\end{table}

\section{Direct Validation of the PRM}
\label{app:prm_validation}

We complement the indirect PRM evaluation in
\Cref{sec:prm-reranking} with direct validation of label quality
using both human and frontier-model audits.

\paragraph{Human audit.}
Our research team manually evaluated a sample of 50 trajectories using
the same annotation criteria as the QwQ-32B labeler. The resulting
step-level annotations showed 84\% agreement with QwQ-32B, providing
an independent manual validation of the generated supervision.

\paragraph{Frontier-model audit.}
We further audit 500 stratified samples drawn from QwQ-32B-labeled
trajectories (held-out from PRM training) by having an independent
frontier-class LLM (Claude Opus 4.7) re-evaluate them under the
\emph{identical} labeling rubric (R1--R6,
\Cref{app:labeling_rubric}).

\begin{table}[!ht]
\centering
\captionsetup{justification=raggedright,singlelinecheck=false}
\caption{Agreement between QwQ-32B (our annotator) and Claude Opus 4.7
on the 500-sample audit.}
\label{tab:prm_validation}
\small
\begin{tabular}{@{}lp{0.46\columnwidth}@{}}
\toprule
\textbf{Metric} & \textbf{Value} \\
\midrule
Audited samples       & 500 \\
Agreement             & \textbf{80.56\%} (402 / 499 with valid labels)\\
Cohen's $\kappa$      & \textbf{0.6104} (substantial) \\
QwQ-labeled GOOD      & 274 \\
Frontier-labeled GOOD & 255 \\
\bottomrule
\end{tabular}
\end{table}

\begin{table}[!ht]
\centering
\caption{Per-step-type agreement on the 500-sample audit.}
\label{tab:prm_validation_per_type}
\small
\begin{tabular}{@{}lrrr@{}}
\toprule
\textbf{Step type} & \textbf{N} & \textbf{Agreement} & \textbf{Cohen's $\kappa$} \\
\midrule
answer & 200 & 91.00\% & \textbf{0.82} \\
search & 249 & 77.11\% & 0.52 \\
reason & 50  & 56.00\% & 0.12 \\
\bottomrule
\end{tabular}
\end{table}

\paragraph{Findings.}
\label{app:annotator_cost}
$\kappa = 0.61$ falls in the \emph{substantial} agreement range
under the standard Landis--Koch \cite{landis1977measurement} interpretation
($\kappa \in [0.61, 0.80]$ = substantial). The disagreement direction
is mild (frontier slightly stricter, $\sim$7\% fewer GOOD calls), with
no systematic permissive bias from QwQ. Per step type, agreement is almost perfect on answer steps ($\kappa = 0.82$) and moderate on
search steps ($\kappa = 0.52$); the lower agreement on intermediate
reasoning steps ($\kappa = 0.12$) reflects the inherent subjectivity of
judging whether a $\langle$think$\rangle$ block provides
``new useful information'' --- but reasoning steps account for only
10\% of the audit, and the labels that drive PRM signal most directly
(answer correctness and search quality) are precisely where frontier
agreement is strongest. This confirms QwQ-32B as a trustworthy open-source labeler.

\section{Statistical Significance of Main Results}
\label{app:significance}

We supplement Table~\ref{tab:main_results} with bootstrap confidence intervals and per-dataset paired $t$-tests, computed on per-question results against \textsc{PRO-Step}. All evaluations are conducted with temperature 0 (greedy decoding), which removes sampling stochasticity and yields deterministic per-question outputs. We therefore report significance over the per-question score distribution rather than across multiple seeds.

\begin{table}[!ht]
\centering
\caption{Macro-AVG bootstrap 95\% confidence intervals ($B=10^4$) for
$\Delta$ vs.\ each prior baseline. Significance: $\star\star\star$ if
the 95\% CI excludes 0.}
\label{tab:bootstrap_ci}
\small
\resizebox{\columnwidth}{!}{%
\begin{tabular}{@{}lcc@{}}
\toprule
\textbf{vs.\ Baseline} & \textbf{$\Delta$ EM (95\% CI)} & \textbf{$\Delta$ F1 (95\% CI)} \\
\midrule
Search-R1   & +2.51 [+1.01, +4.06] $\star\star\star$ & +3.37 [+1.94, +4.82] $\star\star\star$ \\
ReasonRAG   & +1.93 [+0.46, +3.36] $\star\star\star$ & +3.14 [+1.56, +4.65] $\star\star\star$ \\
StepSearch  & +1.36 [-0.29, +3.00] \textsc{ns}        & +2.10 [+0.38, +3.77] $\star\star\star$ \\
\bottomrule
\end{tabular}%
}
\end{table}

\begin{table}[!ht]
\centering
\caption{Per-dataset paired $t$-test on EM. $n$ = test split size.}
\label{tab:per_dataset_t}
\small
\resizebox{\columnwidth}{!}{%
\begin{tabular}{@{}lrccc@{}}
\toprule
\textbf{Dataset} & $n$ & \textbf{vs.\ Search-R1} & \textbf{vs.\ ReasonRAG} & \textbf{vs.\ StepSearch} \\
\midrule
HotpotQA   &  7{,}405 & +0.85, $p$=0.10 \textsc{ns}       & +2.36, $p<10^{-6}$  $\star\star\star$ & +0.01, $p$=0.98 \textsc{ns} \\
PopQA      & 14{,}267 & -0.18, $p$=0.56 \textsc{ns}       & +2.69, $p<10^{-15}$ $\star\star\star$ & +1.23, $p<10^{-3}$ $\star\star\star$ \\
\textbf{2WikiMultiHopQA} & 12{,}576 & \textbf{+9.20}, $p<10^{-75}$ $\star\star\star$ & +4.27, $p<10^{-18}$ $\star\star\star$ & +3.69, $p<10^{-13}$ $\star\star\star$ \\
Bamboogle  &     125 & +3.20, $p$=0.40 \textsc{ns}        & -1.60, $p$=0.66 \textsc{ns}            & +3.20, $p$=0.43 \textsc{ns} \\
MuSiQue    &  2{,}417 & -0.50, $p$=0.46 \textsc{ns}        & +1.90, $p$=0.004 $\star\star$           & -1.32, $p$=0.054 \textsc{ns} \\
\bottomrule
\end{tabular}}
\end{table}

\paragraph{Findings.}
\textsc{PRO-Step} significantly outperforms Search-R1 and ReasonRAG in macro-AVG EM and F1 (95\% CI). Against StepSearch, F1 improvements are significant; while the EM CI marginally includes 0, \textsc{PRO-Step} leads significantly on PopQA and 2WikiMultiHopQA. The decisive 2WikiMultiHopQA lead ($p < 10^{-75}$) indicates the gap is not an averaging artifact.

\section{Generalization Across Model Families, Scales, and Alignments}
\label{app:backbone}

The main generalization results are reported in
\Cref{sec:generalization}. Here, we provide additional analysis of the
3B setting that is not included in the main text, including the
underlying base-model performance, statistical significance, and
training details.

\paragraph{Additional analysis at the 3B scale.}
Table~\ref{tab:results_3b} reports the complete 3B results, including
the underlying Qwen2.5-3B Base model. \textsc{PRO-Step} achieves
31.04 / 40.00 average EM / F1 compared with 26.22 / 33.51 for
Search-R1 3B, while using 5k seed questions compared with Search-R1's 170k
training questions.

\begin{table}[t]
\centering
\caption{Results at the 3B scale (EM / F1, \%).
\textbf{Bold} denotes the best result; ``*'' denotes
$p<0.05$ against Search-R1.}
\label{tab:results_3b}
\footnotesize
\setlength{\tabcolsep}{3.5pt}
\renewcommand{\arraystretch}{1.08}
\resizebox{\columnwidth}{!}{%
\begin{tabular}{@{}lccc@{}}
\toprule
\textbf{Dataset}
& \textbf{Base 3B}
& \textbf{Search-R1}
& \textbf{PRO-Step} \\
\midrule
HotpotQA
& 5.31 / 17.36
& 31.82 / 41.30
& \textbf{32.82 / 44.33*} \\

PopQA
& 7.24 / 22.23
& 36.87 / 42.14
& \textbf{39.88* / 46.10*} \\

2WikiMulti
& 2.27 / 15.28
& 35.96 / 41.86
& \textbf{44.10* / 50.77*} \\

Bamboogle
& 4.00 / 10.64
& 18.40 / 27.97
& \textbf{28.00* / 38.96*} \\

MuSiQue
& 1.24 / 6.27
& 8.07 / 14.27
& \textbf{10.38* / 19.85*} \\

\midrule
\textbf{Average}
& 4.01 / 14.35
& 26.22 / 33.51
& \textbf{31.04 / 40.00} \\
\bottomrule
\end{tabular}}
\end{table}

\paragraph{Training at the 3B scale.}
Direct DPO on the 3B base model achieves only 12.40 EM.
We therefore use SFT warmup to learn the interaction format before
preference optimization. The successful 3B configuration additionally
uses trajectories generated by the larger policy model.

\section{Intermediate-Flawed-But-Correct Analysis}
\label{app:ifbc}
We provide a direct quantitative analysis of the failure mode targeted
by \textsc{PRO-Step}: trajectories that contain flawed intermediate
steps while nevertheless producing the correct final answer.

\paragraph{Definition.}
We define an \emph{intermediate-flawed-but-correct} (IFBC) trajectory
as a trajectory containing at least one flawed intermediate step
(all steps except the final-answer step) while still producing a
correct final answer. We report

\begin{equation}
\mathrm{IFBC}
= P(\text{flaw} \geq 1 \mid \text{correct}).
\end{equation}

Here, flaw denotes at least one flawed intermediate step, and correct denotes a correct final answer.
Lower IFBC therefore indicates that correct answers are less frequently
supported by flawed intermediate trajectories. Steps are annotated
using QwQ-32B under the same R1--R6 rubric used for PRM supervision,
with the annotator blind to system identity. The analysis uses the
trajectories underlying the main experiments for 500 paired questions
each from HotpotQA, 2WikiMultiHopQA, and MuSiQue.

\begin{table}[!ht]
\centering
\caption{Intermediate-Flawed-But-Correct (IFBC) rate among correct
trajectories. Statistical tests compare each method with
\textsc{PRO-Step}. Lower is better.}
\label{tab:ifbc}
\small
\resizebox{\columnwidth}{!}{%
\begin{tabular}{@{}lccc@{}}
\toprule
\textbf{Method}
& \textbf{IFBC $\downarrow$}
& \textbf{$p$ vs.\ Ours}
& \textbf{\# Correct} \\
\midrule

\textsc{PRO-Step}
& 26.8\%
& --
& \textbf{474} \\

Outcome-only ablation
& 40.3\%
& $<10^{-4}$
& 447 \\

Search-R1
& 33.3\%
& 0.033
& 439 \\

StepSearch
& 33.0\%
& 0.037
& 457 \\

ReasonRAG
& \textbf{21.6\%}
& 0.071 (\textsc{ns})
& 430 \\

\bottomrule
\end{tabular}}
\end{table}

\paragraph{Results.}
The controlled comparison with the outcome-only ablation isolates the
effect of process-level preference construction: both variants use the
same training trees and DPO hyperparameters, while differing in how
preference pairs are selected. PRM-guided selection reduces IFBC from
40.3\% to 26.8\%, a reduction of 13.5 percentage points
($p < 10^{-4}$), while increasing the number of correct trajectories
from 447 to 474.

The same pattern appears in the training signal. Outcome-only
selection promotes 19.8\% flawed-but-successful trajectories to the
chosen side of the DPO pairs, whereas PRM-guided selection reduces this
rate to 10.7\%.

\textsc{PRO-Step} also exhibits significantly lower IFBC rates than
Search-R1 and StepSearch. ReasonRAG has a numerically lower IFBC rate,
but the difference is not statistically significant ($p=0.071$) and
it produces fewer correct trajectories (430 vs.\ 474). Thus, the IFBC
analysis is interpreted jointly with task performance rather than as
a standalone measure of system quality.

\section{Detailed Recovery Analysis}
\label{app:recovery}

We provide the full breakdowns underlying \Cref{sec:recovery}. A
retrieval succeeds when the gold passage appears among the top-3
retrieved documents, and recovery depth $k$ denotes the first
successful retrieval step, with the initial retrieval indexed as
step 0. Throughout, $n$ is reported as Search-R1\,/\,PRO-STEP, since the
recovered subset is defined by each model's own search behavior.

\Cref{tab:recovery_f1} reports token-level F1 on the same recovered
subsets as \Cref{tab:recovery}. \textsc{PRO-Step} leads on four of
five datasets, with the largest margin on 2WikiMulti (+7.8). The
exception is Bamboogle, whose recovered subset contains only 25
questions.

\Cref{tab:recovery_depth} breaks 2WikiMulti down by recovery depth.
Gains are consistent at $k=1$ through $k=3$. The $k\geq4$ bucket
favors \textsc{PRO-Step} by a wide margin, but both subsets are small
and the comparison is not reliable.

\Cref{tab:no_recovery} reports the complementary case in which the
initial retrieval already surfaces the gold passage, so no recovery
is required. \textsc{PRO-Step} performs competitively or better on
four of five datasets, indicating that its advantage on recovered
trajectories does not stem from an easier subset.

\begin{table}[t]
\centering
\caption{Token-level F1 on recovered trajectories.}
\label{tab:recovery_f1}
\small
\renewcommand{\arraystretch}{1.10}
\begin{tabular*}{\columnwidth}{@{\extracolsep{\fill}}lccc@{}}
\toprule
\textbf{Dataset} & \textbf{Search-R1} & \textbf{\textsc{PRO-Step}} & \textbf{$\Delta$} \\
\midrule
HotpotQA   & 66.6 & \textbf{69.0} & +2.5 \\
PopQA      & 54.2 & \textbf{57.6} & +3.4 \\
2WikiMulti & 58.2 & \textbf{66.1} & +7.8 \\
Bamboogle  & \textbf{87.6} & 84.0 & $-$3.6 \\
MuSiQue    & 48.9 & \textbf{51.1} & +2.1 \\
\bottomrule
\end{tabular*}
\end{table}

\begin{table}[t]
\centering
\caption{Final EM on 2WikiMulti by recovery depth $k$.}
\label{tab:recovery_depth}
\small
\renewcommand{\arraystretch}{1.10}
\begin{tabular*}{\columnwidth}{@{\extracolsep{\fill}}lcccc@{}}
\toprule
\textbf{Method} & \textbf{$k=1$} & \textbf{$k=2$} & \textbf{$k=3$} & \textbf{$k\geq4$} \\
\midrule
$n$ & 3{,}193 / 2{,}565 & 578 / 332 & 150 / 33 & 81 / 15 \\
\midrule
Search-R1 & 49.0 & 42.6 & 38.7 & 34.6 \\
\textsc{PRO-Step} & \textbf{53.1} & \textbf{50.9} & \textbf{45.5} & \textbf{60.0} \\
\midrule
$\Delta$ & +4.1 & +8.3 & +6.8 & +25.4 \\
\bottomrule
\end{tabular*}
\end{table}

\begin{table}[t]
\centering
\caption{Final EM when the initial retrieval already succeeds.
Subset sizes ($n$, ours\,/\,Search-R1) are 3{,}136 / 3{,}434,\;
8{,}597 / 8{,}776,\;
4{,}302 / 4{,}732,\;
23 / 25, and
348\,/\,405, in row order.}
\label{tab:no_recovery}
\small
\renewcommand{\arraystretch}{1.10}
\begin{tabular*}{\columnwidth}{@{\extracolsep{\fill}}lccc@{}}
\toprule
\textbf{Dataset} & \textbf{Search-R1} & \textbf{\textsc{PRO-Step}} & \textbf{$\Delta$} \\
\midrule
HotpotQA   & 53.1 & \textbf{54.1} & +1.0 \\
PopQA      & \textbf{62.3} & 61.0 & $-$1.3 \\
2WikiMulti & 42.5 & \textbf{54.6} & +12.1 \\
Bamboogle  & 34.8 & \textbf{64.0} & +29.2 \\
MuSiQue    & 25.3 & \textbf{27.2} & +1.9 \\
\bottomrule
\end{tabular*}
\end{table}

\section{Case Study}
\label{app:case-study}

We present qualitative examples to illustrate the limitation of
outcome-based learning: even when generating reasoning step-by-step,
models trained solely on final answers fail to penalize intermediate
logical gaps. In contrast, \textsc{PRO-Step}'s PRM evaluates the
validity of every single step, effectively preventing early mistakes
from ruining the entire trajectory.

\begin{figure*}[ht]
\centering
\includegraphics[width=\textwidth]{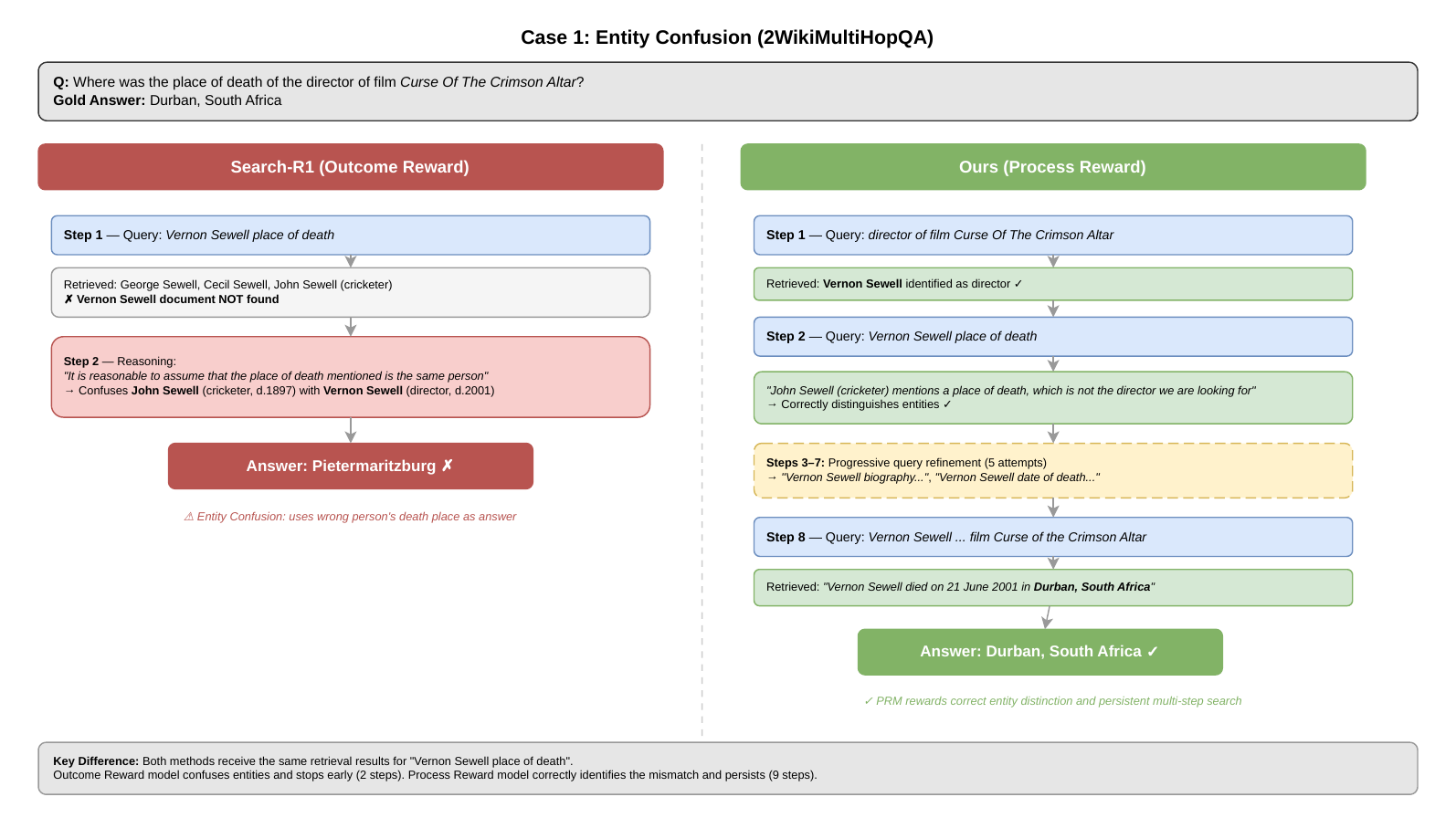}
\caption{Case Study 1: Entity confusion. The baseline prematurely
answers based on a mismatched entity (John Sewell), whereas
\textsc{PRO-Step} rejects the flawed step via PRM feedback and
persistently searches for the correct target.}
\label{fig:case-study-1}
\end{figure*}

\paragraph{Case Study 1: Resolving Entity Confusion.}
As shown in \Cref{fig:case-study-1}, the baseline confuses the target
``Vernon Sewell'' with ``John Sewell.'' Blind to this semantic drift,
it prematurely outputs an incorrect answer at Step 2. Conversely,
\textsc{PRO-Step}'s PRM immediately detects this entity mismatch and
assigns a BAD signal. Forced to reject the spurious path, the model
continues exploring alternative queries until it retrieves the correct
evidence at Step 8 and answers correctly at Step 9.

\begin{figure*}[ht]
\centering
\includegraphics[width=\textwidth]{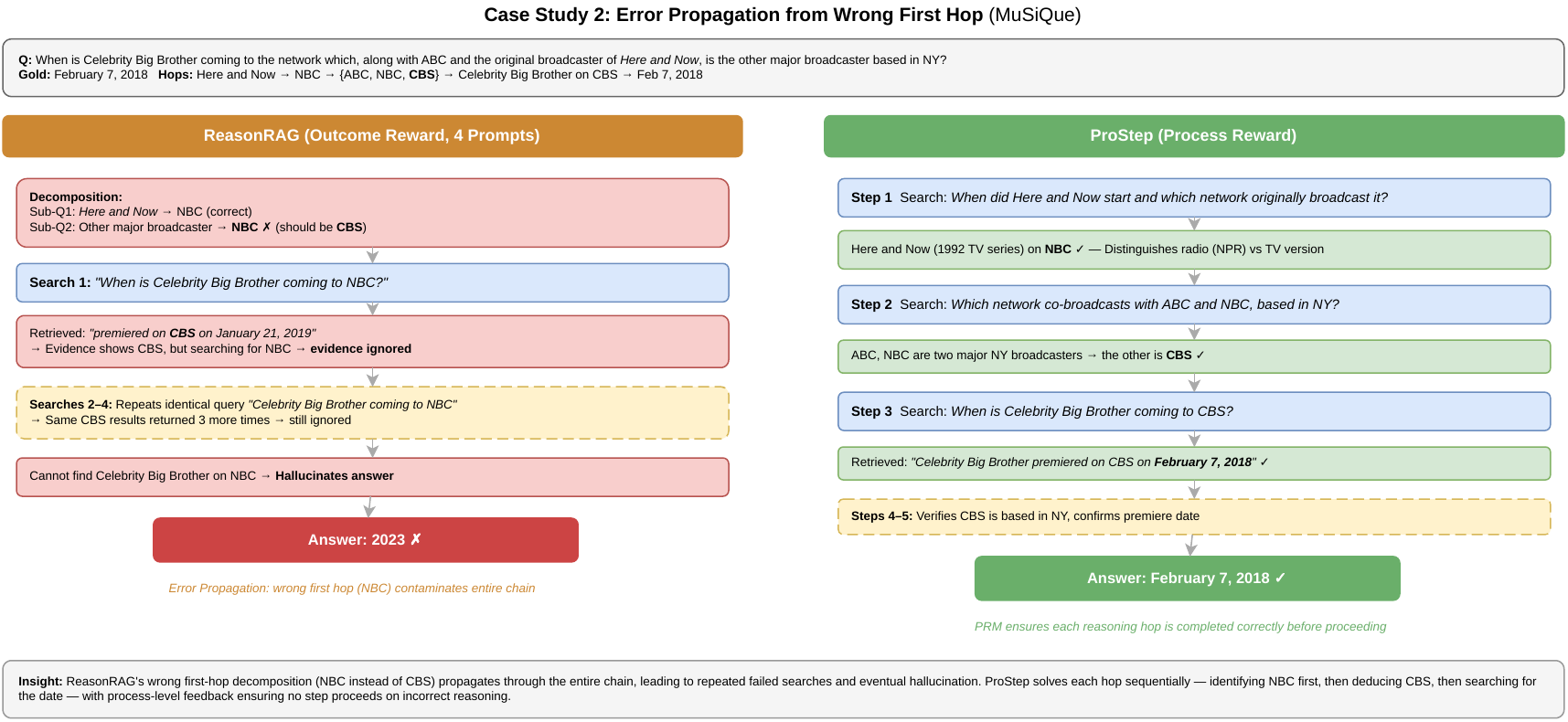}
\caption{Case Study 2: Error propagation. The baseline repeats useless
searches after an incorrect intermediate hop. PRO-Step evaluates each hop,
avoids the error made by the baseline, and completes the logical chain.}
\label{fig:case-study-2}
\end{figure*}

\paragraph{Case Study 2: Preventing Error Propagation.}
\Cref{fig:case-study-2} demonstrates the danger of error propagation
in multi-hop queries. The baseline (ReasonRAG) incorrectly decomposes
an intermediate hop (e.g., targeting ``NBC'' instead of ``CBS''). Without
step-level correction, it stubbornly repeats the same futile search
four times before hallucinating. \textsc{PRO-Step}, however, receives
step-level PRM verification. By recognizing the invalid intermediate
conclusion early, it pivots its strategy and successfully constructs
the correct logical chain (NBC $\rightarrow$ CBS $\rightarrow$ Date).

\section{PRM-Feedback-Guided Regeneration:
A Failed Augmentation Strategy}
\label{app:regen}

To test whether augmenting our preference set with critic-feedback-guided regeneration would improve downstream performance, we ran the following pipeline. For each of the 1{,}456 questions where the policy failed to produce a correct answer (final $F_1 = 0$), we injected the PRM's \texttt{critic\_feedback} from the failure point into the user prompt and had the base Qwen2.5-7B-Instruct regenerate from the last sound prefix. The regenerated trajectories were re-scored by the PRM, and we applied the same outcome-margin filter (chosen $F_1 \geq 0.2$ and $\Delta F_1 \geq 0.2$) as in our main pipeline. This yielded 20{,}567 preference pairs over 3{,}128 questions (compared to 15{,}877 pairs over 2{,}866 questions in the original outcome filter set). We then trained DPO under identical hyperparameters ($\beta = 0.1$, LoRA $r=64/\alpha=128$) on the augmented set.

\paragraph{Training metrics improve while evaluation collapses.}
\Cref{tab:regen_train} shows that the augmented training run yields substantially stronger DPO statistics: the loss roughly halves (0.42 $\rightarrow$ 0.22), the preference margin grows by 60\% (2.17 $\rightarrow$ 3.48), and training accuracy climbs from 81\% to 91\%. However, downstream evaluation collapses on every dataset, with average EM dropping from 34.51 to 23.83 and average F1 from 44.09 to 34.27 (\Cref{tab:regen_eval}).

\begin{table}[ht]
\centering
\small
\caption{Training metrics for the outcome filter main set vs.\ the
regen-augmented set under identical DPO hyperparameters
($\beta=0.1$).}
\label{tab:regen_train}
\begin{tabular*}{\columnwidth}{@{\extracolsep{\fill}}lccc@{}}
\toprule
\textbf{Variant} & \textbf{Loss} & \textbf{Margin} & \textbf{Acc} \\
\midrule
outcome filter main & 0.42 & 2.17 & 81\% \\
\quad + regen & 0.22 & 3.48 & 91\% \\
\bottomrule
\end{tabular*}
\end{table}

\begin{table*}[ht]
\centering
\caption{Downstream evaluation: the regen-augmented variant degrades on every benchmark despite stronger training metrics.}
\label{tab:regen_eval}
\resizebox{\textwidth}{!}{%
\begin{tabular}{@{}lccccccc@{}}
\toprule
\textbf{Variant} & \textbf{PopQA} & \textbf{HotpotQA} & \textbf{2WikiMulti} & \textbf{Bamboogle} & \textbf{MuSiQue} & \textbf{Average} \\
\midrule
outcome filter main & 40.47 / 47.37 & 38.73 / 51.63 & 44.07 / 51.43 & 36.80 / 47.63 & 12.49 / 22.41 & \textbf{34.51 / 44.09} \\
outcome filter + regen & 28.39 / 37.55 & 26.02 / 39.08 & 33.25 / 42.63 & 24.80 / 35.30 & 6.70 / 16.80 & 23.83 / 34.27 \\
$\Delta$ & $-12.08$ / $-9.82$ & $-12.71$ / $-12.55$ & $-10.82$ / $-8.80$ & $-12.00$ / $-12.33$ & $-5.79$ / $-5.61$ & $-10.68$ / $-9.82$ \\
\bottomrule
\end{tabular}%
}
\end{table*}

\paragraph{Diagnosis: action-format mismatch between training and inference.}
The regenerated chosen trajectories were produced by the base policy under explicit critic-feedback prompting, which alters the model's output distribution: the regenerated $\langle$\texttt{think}$\rangle$ blocks become verbose meta-commentary referencing the feedback (e.g., reasoning about why the previous step was flawed), and the action selection becomes biased toward what the feedback suggests rather than what the trajectory state requires. At inference time, the deployed policy operates without any such feedback channel, so the distribution it must produce diverges from the regenerated distribution it was trained to match.

The outcome-margin filter, which enforces $F_1 \geq 0.2$ and $\Delta F_1 \geq 0.2$, can verify answer-level quality but is blind to this stylistic drift in intermediate tokens. As a result, the augmented training set rewards the policy for matching feedback-conditioned outputs that never appear at test time, and the strong DPO metrics reflect overfitting to this artificial distribution rather than genuine improvement in multi-hop reasoning. We view this as a concrete failure mode of naive PRM-rationale-guided augmentation, and a meaningful constraint that future work in this direction must address.

\section{Released Resources}
\label{app:resources}
Our implementation and complete resource index are available in our
\href{https://github.com/keemminnke/PRO-Step}{GitHub repository}.

\paragraph{Models.}
\begin{itemize}
    \item \textbf{Policy Model:}
    \href{https://huggingface.co/MinKeonKim/PRO-STEP-Policy-7B}
    {PRO-STEP-Policy-7B}.
    
    \item \textbf{Process Reward Model:}
    \href{https://huggingface.co/MinKeonKim/PRO-STEP-PRM-8B}
    {PRO-STEP-PRM-8B}.
\end{itemize}

\paragraph{Datasets.}
\begin{itemize}
    \item \textbf{DPO Preference Pairs:}
    \href{https://huggingface.co/datasets/MinKeonKim/PRO-STEP-Preference-Data}
    {PRO-STEP-Preference-Data}.
    
    \item \textbf{PRM Training Annotations:}
    \href{https://huggingface.co/datasets/MinKeonKim/PRO-STEP-PRM-Data}
    {PRO-STEP-PRM-Data}.
\end{itemize}

\end{document}